\documentclass[10pt,journal,compsoc]{IEEEtran}
\usepackage[utf8]{inputenc}
\usepackage[T1]{fontenc}

\usepackage{color,xcolor}
\usepackage{epsfig}
\usepackage{graphicx}

\usepackage{adjustbox}
\usepackage{array}
\usepackage{booktabs}
\usepackage{colortbl}
\usepackage{float,wrapfig}
\usepackage{hhline}
\usepackage{multirow}
\usepackage{subcaption} %
\usepackage[font=scriptsize]{caption}

\usepackage{amsmath,amsfonts,amsthm,amssymb}
\usepackage{bm}
\usepackage{nicefrac}
\usepackage{microtype}

\usepackage{changepage}
\usepackage{extramarks}
\usepackage{fancyhdr}
\usepackage{lastpage}
\usepackage{setspace}
\usepackage{soul}
\usepackage{xspace}
\usepackage{indentfirst}
\usepackage{pifont}
\usepackage{cuted}
\usepackage{wrapfig}

\usepackage[pagebackref=true,breaklinks=true,letterpaper=true,colorlinks,bookmarks=false]{hyperref}
\usepackage{url}

\usepackage{algorithm, algorithmic}
\usepackage{enumitem}

\usepackage{wasysym}
\newcolumntype{L}[1]{>{\raggedright\let\newline\\\arraybackslash\hspace{0pt}}m{#1}}
\newcolumntype{C}[1]{>{\centering\let\newline\\\arraybackslash\hspace{0pt}}m{#1}}
\newcolumntype{R}[1]{>{\raggedleft\let\newline\\\arraybackslash\hspace{0pt}}m{#1}}

\newcommand{\sect}[1]{Section~\ref{sect:#1}}
\newcommand{\app}[1]{Appendix~\ref{app:#1}}

\newcommand{\fig}[1]{Figure~\ref{fig:#1}}
\newcommand{\figs}[1]{Figures~\ref{fig:#1}}
\newcommand{\tab}[1]{Table~\ref{tab:#1}}

\newcommand{\fid}{Fr\'echet Inception Distance\xspace}
\newcommand{\lblfig}[1]{\label{fig:#1}}
\newcommand{\lblsect}[1]{\label{sect:#1}}
\newcommand{\lblapp}[1]{\label{app:#1}}

\newcommand{\lbltab}[1]{\label{tab:#1}}

\newcommand{\ignorethis}[1]{}

\def\naive{na\"{\i}ve\xspace}
\def\Naive{Na\"{\i}ve\xspace}

\makeatletter
\DeclareRobustCommand\onedot{\futurelet\@let@token\@onedot}
\def\@onedot{\ifx\@let@token.\else.\null\fi\xspace}

\def\eg{\emph{e.g}\onedot} 
\def\ie{\emph{i.e}\onedot} 
 
 \def\vs{\emph{vs}\onedot}
 
\def\etal{\emph{et al}\onedot}
\makeatother
\definecolor{citecolor}{RGB}{34,139,34}
\definecolor{mydarkblue}{rgb}{0,0.08,1}
\definecolor{mydarkgreen}{rgb}{0.02,0.6,0.02}
\definecolor{mydarkred}{rgb}{0.8,0.02,0.02}
\definecolor{mydarkorange}{rgb}{0.40,0.2,0.02}
\definecolor{mypurple}{RGB}{111,0,255}
\definecolor{myred}{rgb}{1.0,0.0,0.0}
\definecolor{mygold}{rgb}{0.75,0.6,0.12}
\definecolor{mydarkgray}{rgb}{0.66, 0.66, 0.66}

\newcommand{\myparagraph}[1]{\vspace{5pt}\noindent\textbf{#1}}

\def\method{Spatially Sparse Inference\xspace}
\def\methodabbr{SSI\xspace}
\def\engine{Sparse Incremental Generative Engine\xspace}
\def\progdist{PD\xspace}
\def\stbldiff{SD\xspace}
\def\engineabbr{SIGE\xspace}

\def\multirowcenter{-0.5\dimexpr \aboverulesep + \belowrulesep + \cmidrulewidth}

\newcommand{\act}{\ensuremath{A}}
\newcommand{\actori}{\ensuremath{\act^{\text{original}}}}
\newcommand{\actedi}{\ensuremath{\act^{\text{edited}}}}
\newcommand{\actdel}{\ensuremath{\Delta\act}}
\newcommand{\weight}{\ensuremath{W}}
\newcommand{\bias}{\ensuremath{b}}
\newcommand{\conv}{\ensuremath{F}}

\ifCLASSOPTIONcompsoc
  \usepackage[nocompress]{cite}
\else
  \usepackage{cite}
\fi

\ifCLASSINFOpdf
\else
\fi

\begin{document}

\title{Efficient Spatially Sparse Inference for\\Conditional GANs and Diffusion Models}

\author{
	Muyang Li, Ji Lin, Chenlin Meng, Stefano Ermon, Song Han and Jun-Yan Zhu
	\IEEEcompsocitemizethanks{
		\IEEEcompsocthanksitem M. Li, J. Lin and S. Han are with Massachusetts Institute of Technology.
		\IEEEcompsocthanksitem C. Meng and S. Ermon are with Stanford University.
		\IEEEcompsocthanksitem J.-Y. Zhu is with Carnegie Mellon University. 
	}
}

\markboth{IEEE TRANSACTIONS ON PATTERN ANALYSIS AND MACHINE INTELLIGENCE, VOL. X, NO. X, MMMMMMM YYYY}
{Li \MakeLowercase{\textit{et al.}}: Efficient Spatially Sparse Inference for Conditional GANs and Diffusion Models}

\IEEEtitleabstractindextext{
	\begin{abstract}
		During image editing, existing deep generative models tend to re-synthesize the entire output from scratch, including the unedited regions. This leads to a significant waste of computation, especially for minor editing operations. In this work, we present \method (\methodabbr), a general-purpose technique that selectively performs computation for edited regions and accelerates various generative models, including both conditional GANs and diffusion models. Our key observation is that users prone to gradually edit the input image. This motivates us to cache and reuse the feature maps of the original image. Given an edited image, we sparsely apply the convolutional filters to the edited regions while reusing the cached features for the unedited areas. Based on our algorithm, we further propose \engine (\engineabbr) to convert the computation reduction to latency reduction on off-the-shelf hardware. With about $1\%$-area edits, SIGE accelerates DDPM by $3.0\times$ on NVIDIA RTX 3090 and $4.6\times$ on Apple M1 Pro GPU, Stable Diffusion by $7.2\times$ on 3090, and GauGAN by $5.6\times$ on 3090 and $5.2\times$ on M1 Pro GPU. Compared to our conference version, we extend \engineabbr to accommodate attention layers and apply it to Stable Diffusion. Additionally, we offer support for Apple M1 Pro GPU and include more results with large and sequential edits.
	\end{abstract}

	\begin{IEEEkeywords}
		Diffusion Models, GAN, Sparse, Image Editing, Efficiency
	\end{IEEEkeywords}
}

\maketitle

\IEEEdisplaynontitleabstractindextext

\IEEEpeerreviewmaketitle

\begin{strip}
    \centering
    \vspace{-85pt}
    \includegraphics[width=\linewidth]{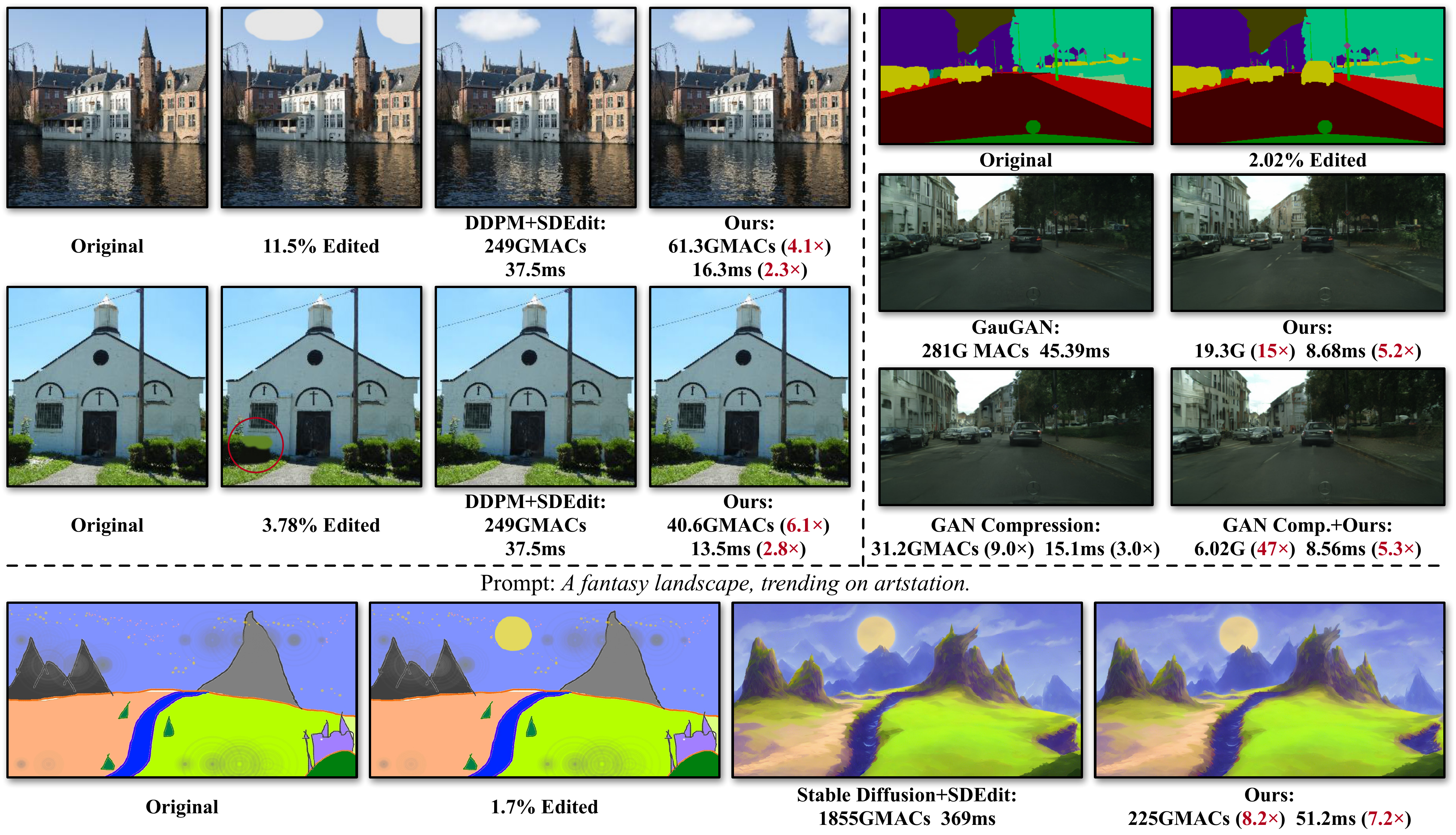}
    \vspace{-20pt}
    \captionof{figure}{
    	We introduce \textit{\engine (\engineabbr)}, an engine that selectively performs computations at the edited regions for image editing applications. The computation and latency are measured on NVIDIA RTX 3090 for a single forward. For the above examples, \engineabbr significantly reduces the computation of SDEdit with DDPM~\cite{song2020denoising,ho2020denoising} and Stable Diffusion~\cite{rombach2022high}, and GauGAN~\cite{park2019semantic} while preserving the image quality. When combined with existing model compression methods such as GAN Compression~\cite{li2020gan}, it further reduces the computation of GauGAN by $47\times$. 
    }
    \vspace{8pt}
    \lblfig{teaser}
\end{strip}
\ifCLASSOPTIONcompsoc
\IEEEraisesectionheading{\section{Introduction}\label{sec:introduction}}
\else
\section{Introduction}
\lblsect{Introduction}
\fi
\begin{figure*}[t]
    \centering
    \includegraphics[width=\linewidth]{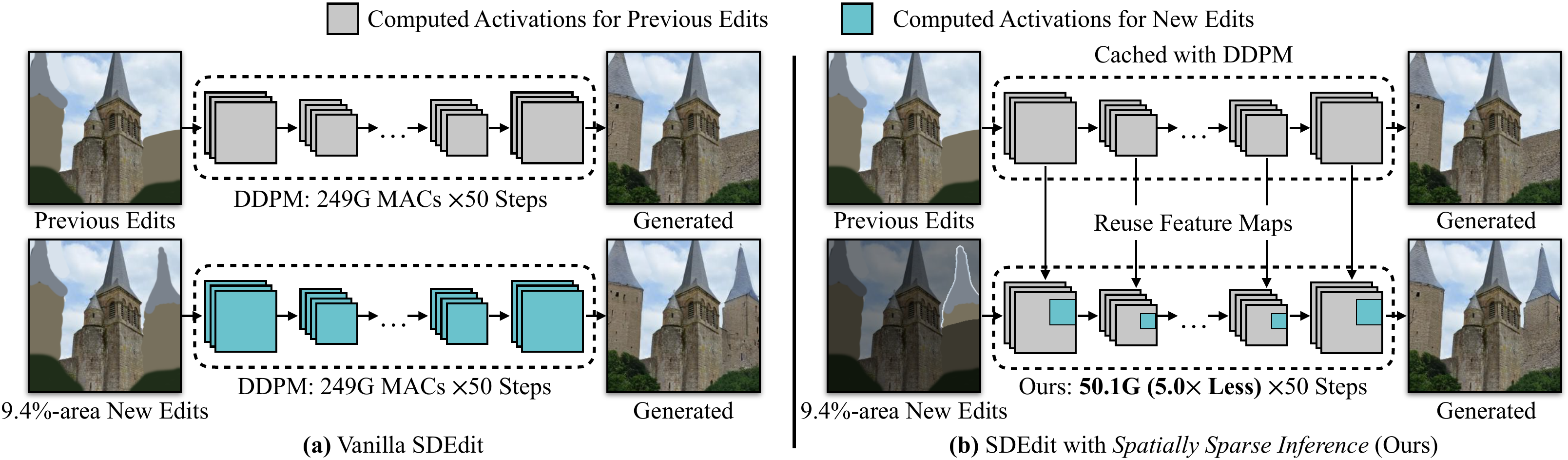}
    \vspace{-15pt}
	\caption{
		\looseness=-1 
		In the interactive editing scenario, a user adds a new building, which occupies 9.4\% pixels. \textbf{(a)} Vanilla SDEdit has to apply denoising networks to the \textit{entire} image, even though only a 9.4\% area was edited. \textbf{(b)} Our method instead \textit{reuses} the feature maps of the previous edits and only sparsely applies convolutions to the \textit{newly edited} regions, which results in a $5.0\times$ MACs reduction for this example.
	}
	\vspace{-15pt}
    \lblfig{idea}
\end{figure*}

\IEEEPARstart{D}{eep} generative models, such as GANs~\cite{goodfellow2014generative,karras2019style} and diffusion models~\cite{sohl2015deep,ho2020denoising,song2020denoising}, excel at synthesizing photo-realistic images, enabling many image synthesis and editing applications. For example, users can edit an image by drawing sketches~\cite{isola2017image,sangkloy2017scribbler}, semantic maps~\cite{isola2017image,park2019semantic}, or strokes~\cite{meng2022sdedit}. All of these applications require users to interact with generative models frequently and therefore demand short inference time. 

\looseness=-1
In practice, content creators often edit images gradually and only update a small image region each time. However, even for a minor edit, recent generative models often synthesize the entire image, including the unchanged areas, which leads to a significant waste of computation. As a concrete example shown in \fig{idea}(a), the result of the previous edits has already been computed, and the user further edits 9.4\% areas. However, vanilla SDEdit~\cite{meng2022sdedit} needs to apply the denoising network on the entire image to obtain the newly edited regions, wasting 80\% computation on the unchanged areas. A naive approach to address this issue would be to first segment the newly edited regions, synthesize the corresponding output regions, and blend the outputs back into the previous output. Unfortunately, this method often creates visible seams between the newly edited and unedited regions. How could we save the computation by only updating the edited regions without losing global coherence?

\looseness=-1
In this work, we propose \method (\methodabbr), a general method to accelerate deep generative models, including conditional GANs and diffusion models, by utilizing the spatial sparsity of edited regions. Our method is motivated by the observation that feature maps in the unedited regions remain mostly the same during user editing. As shown in \fig{idea}(b), our key idea is to reuse the cached feature maps of the previous edits and sparsely update the newly edited areas. Specifically, given user input, we first compute a difference mask to locate the newly edited regions. For each convolution layer in the model, we only sparsely apply the filters to the masked regions while reusing the previous activations for the unchanged areas. The sparse update can significantly reduce the computation without hurting the image quality. However, the sparse update involves a gather-scatter process and often incurs significant latency overheads with existing deep learning frameworks. To address the issue, we propose \textit{\engine (\engineabbr)} to translate the theoretical computation reduction of our algorithm to measured latency reduction on various hardware.

\looseness=-1
To evaluate our method, we curate image  editing and inpainting benchmarks on LSUN Church~\cite{yu15lsun}, Cityscapes~\cite{cordts2016cityscapes} and LAION-5B~\cite{schuhmann2022laion}. Without loss of visual fidelity, we reduce the computation of DDPM~\cite{song2020denoising,ho2020denoising}, Progressive Distillation~\cite{salimans2021progressive}, Stable Diffusion~\cite{rombach2022high}, and GauGAN~\cite{park2019semantic} by up to $7.5\times$, $2.7\times$, $8.2\times$, and $18\times$, respectively, measured by MACs\footnote{We measure the computational cost with the number of Multiply-Accumulate operations (MACs). 1 MAC=2 FLOPs.}. Compared to existing generative model acceleration methods~\cite{li2020gan,hou2021slimmable,fu2020autogan,li2022learning,jin2021teachers,shaham2021spatially,wang2020gan}, our method directly uses the off-the-shelf pre-trained weights and could be applied to these methods as a plugin. When applied to GAN Compression~\cite{li2020gan}, we reduce the computation of GauGAN by up $50\times$. See \fig{teaser} for some examples of our method. With \engineabbr, we accelerate DDPM by up to $3.0\times$ on NVIDIA RTX 3090, $4.6\times$ on Apple M1 Pro GPU, and $6.6\times$ on M1 Pro CPU, Stable Diffusion by up to $7.2\times$ on 3090, and GauGAN by up to $5.6\times$ on 3090, $5.2\times$ on M1 Pro GPU, and $14\times$ on M1 Pro CPU. 

\looseness=-1
This journal paper extends our conference version~\cite{li2022efficient} with new development and experiments in the following areas:
\begin{itemize}[leftmargin=*]
	\item \looseness=-1 We extend \engineabbr to support both self-attention and cross-attention layers by pruning unedited query tokens. This optimization can dramatically shrink the attention map size according to the edit size, reducing the computation and latency of the attention layers correspondingly.
    \item We further apply our method to Stable Diffusion~\cite{rombach2022high}, a widely-used text-to-image model with latency primarily bottlenecked by its self-attention layers.  Since our previous engine can only accelerate convolutions, it can only reduce the computation by $1.6\times$ and latency by $1.1\times$ on Stable Diffusion, even with a $2.8\%$-area edit. In contrast, with our new attention optimization, we reduce Stable Diffusion's computation and latency by $\sim 5\times$.
    \item \looseness=-1 We additionally support \engineabbr on the Metal Performance Shaders (MPS) backend to enable inference on the Apple M1 Pro GPU. On this hardware, we achieve up to $4.6\times$, $3.0\times$, and $5.2\times$ speedups for DDPM, Progressive Distillation, and GauGAN, respectively.
	\item We show additional results of \engineabbr with large and sequential edits. Specifically, on NVIDIA RTX 3090, our method remains faster than the original model with up to $\sim 70\%$ edits. For sequential edits, it can incrementally update the cached activations. We also include an extra ablation study regarding the dilation hyper-parameter to validate our design choice.
\end{itemize}

Our code, benchmarks and demo are available at \url{https://github.com/lmxyy/sige}.

\section{Related Work}

\begin{figure*}[t]
    \centering
    \includegraphics[width=\linewidth]{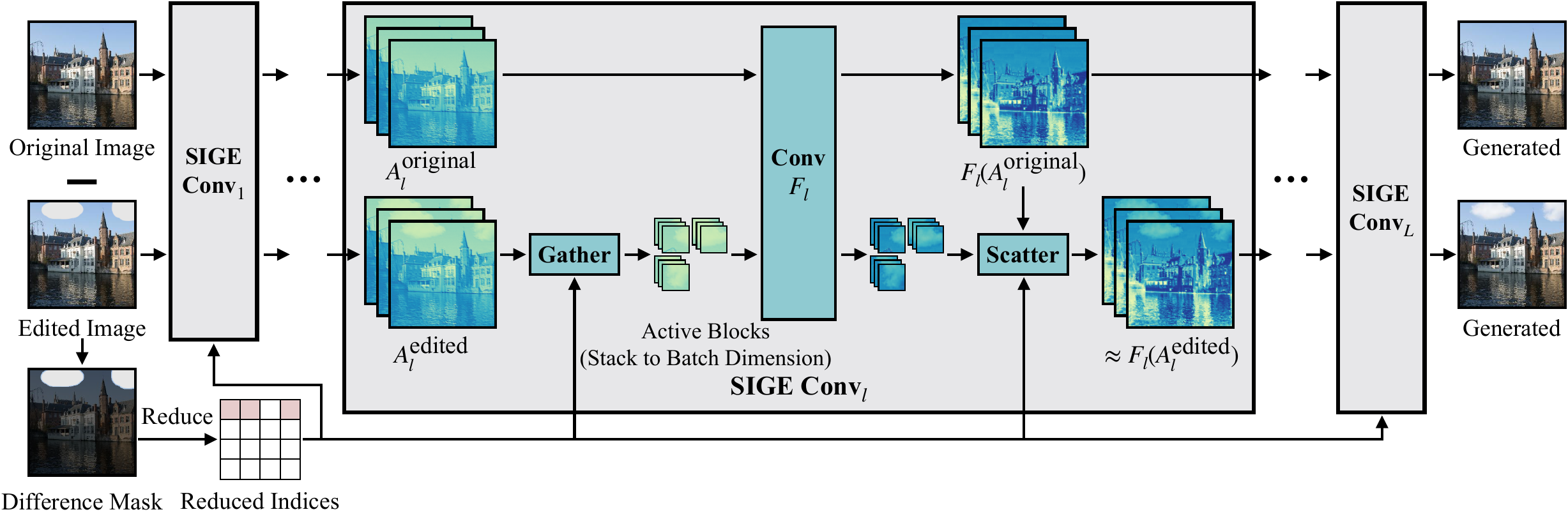}
    \vspace{-15pt}
    \caption{
    Tiling-based sparse convolution overview. For each convolution $\conv_l$ in the network, we wrap it into SIGE Conv$_l$. The activations of the original image are already pre-computed. When getting the edited image, we first compute a difference mask between the original and edited image and reduce the mask to the active block indices to locate the edited regions. In each SIGE Conv$_l$, we directly gather the active blocks from the edited activation $\actedi_l$ according to the reduced indices, stack the blocks along the batch dimension, and feed them into $\conv_l$. The gathered blocks have an overlap of width 2 if $\conv_l$ is $3\times 3$ convolution with stride 1~\cite{ren2018sbnet}. After getting the output blocks from $\conv_l$, we scatter them back into $\conv_l(\actori_l)$ to get the edited output, which approximates $\conv_l(\actedi_l)$.
    }
    \vspace{-10pt}
    \lblfig{method}
\end{figure*} 
\begin{figure*}[t]
    \centering
    \includegraphics[width=\linewidth]{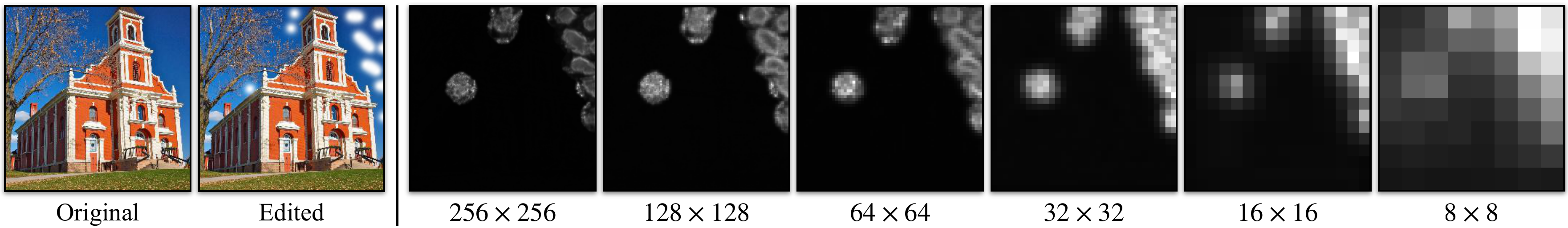}
    \vspace{-15pt}
    \caption{\looseness=-1 Left: Detailed edit example. Right: Channel-wise average of $|\actdel_l|$ at the $l$-th layer of DDPM with different feature map resolutions. $|\actdel_l|$ is sparse and non-zero values are aggregated at the edited regions.}
    \vspace{-15pt}
    \lblfig{activation-difference}
\end{figure*}

\looseness=-1
\myparagraph{Generative models.}
Generative models such as GANs~\cite{goodfellow2014generative,karras2019style,karras2020analyzing,brock2018large}, diffusion models~\cite{ho2020denoising,sohl2015deep,dhariwal2021diffusion,rombach2022high}, and auto-regressive models~\cite{esser2021taming,razavi2019generating} have demonstrated impressive photorealistic synthesis capability. They have also been extended to conditional image synthesis tasks such as image-to-image translation~\cite{saharia2021palette,isola2017image,zhu2017unpaired,zhu2020sean}, controllable image generation~\cite{meng2022sdedit,nichol2021glide,park2019semantic}, and real image editing~\cite{choi2021ilvr, nichol2021glide,kim2021diffusionclip,zhu2016generative,patashnik2021styleclip,abdal2019image2stylegan,abdal2020image2stylegan++,zhu2020sean}. Unfortunately, recent generative models have become increasingly computationally intensive, compared to their recognition counterparts. For example, GauGAN~\cite{park2019semantic} consumes 281GMACs, 500$\times$ more than  MobileNet~\cite{howard2019searching,howard2017mobilenets,sandler2018mobilenetv2}. Similarly, one key limitation of diffusion models~\cite{ho2020denoising} is their substantial computation cost and long inference time. To generate one image, DDPM requires hundreds or thousands of forwarding steps~\cite{ho2020denoising, dhariwal2021diffusion}, which is often infeasible in real-world interactive settings. To improve the sampling efficiency of DDPMs, recent works~\cite{song2020denoising,song2020score,kong2021fast} propose to interpret the sampling process of DDPMs from the perspective of ordinary differential equations. However, these approaches still require hundreds of steps to generate high-quality samples. To further reduce the sampling cost, DDGAN~\cite{xiao2022DDGAN} uses a multimodal conditional GAN to model each denoising step. Salimans \etal.~\cite{salimans2021progressive} propose to progressively distill a pre-trained DDPM model into a new one that requires fewer steps. Although this approach drastically reduces the sampling steps, the distilled model itself remains computationally prohibitive. Unlike prior work, our work focuses on reducing the computation cost of a pre-trained model. It is complementary to recent efforts on model compression, distillation, and the sampling step reduction of the diffusion models.

\looseness=-1 
\myparagraph{Model acceleration.} 
People apply model compression techniques, including pruning~\cite{han2016deep,he2018amc,lin2017runtime,he2017channel,liu2017learning,liu2019metapruning} and quantization~\cite{han2016deep,zhou2016dorefa, rastegari2016xnor,wang2019haq,choi2018pact,jacob2018quantization}, to reduce the computation and model size of off-the-shelf deep learning models. Recent works apply Neural Architecture Search (NAS)~\cite{zoph2017neural,zoph2018learning,liu2019darts,cai2019proxylessnas,tan2019mnasnet,wu2019fbnet,lin2020mcunet} to automatically design efficient neural architectures. The above ideas can be successfully applied to accelerate the inference of GANs~\cite{li2020gan,lin2021anycost,shu2019co,liu2021content,hou2021slimmable,ma2021cpgan,fu2020autogan,li2022learning,jin2021teachers,shaham2021spatially,wang2020gan,aguinaldo2019compressing}. Although these methods have achieved prominent compression and speedup ratios, they all reduce the computation from the model dimension but fail to exploit the redundancy in the spatial dimension during image editing. Besides, these methods require re-training the compressed model to maintain performance, while our method can be directly applied to existing pre-trained models. We show that our method can be combined with model compression~\cite{li2020gan} to achieve a $\sim 50\times$ MACs reduction in \sect{Main Results}.

\looseness=-1 
\myparagraph{Sparse computation.}
Sparse computation has been widely explored in the weight domain~\cite{han2015learning,li2016pruning,liu2015sparse,jaderberg2014speeding}, input domain~\cite{tang2022torchsparse,riegler2017octnet}, and activation domain~\cite{ren2018sbnet,judd2017cnvlutin2,shi2017speeding,dong2017more}. For activation sparsity, RRN~\cite{pan2018recurrent} utilizes the sparsity in the consecutive video frame difference to accelerate video models. However, their sparsity is unstructured, which requires special hardware to reach its full speedup potential. Several works instead use structured sparsity. Li \etal~\cite{li2017not} use a deep layer cascade to apply more convolution layers on the hard regions than the easy regions to improve the accuracy and speed of semantic segmentation. To accelerate 3D object detection,  SBNet~\cite{ren2018sbnet} uses a spatial mask, either from priori problem knowledge or an auxiliary network, to sparsify the activations. It adopts a tiling-based sparse convolution algorithm to handle spatial sparsity. Recent works further integrate the spatial mask generation network into the sparse inference network in an end-to-end manner~\cite{verelst2020dynamic} and extend the idea to different tasks~\cite{wang2021exploring,han2021spatially,wang2022adafocus,parger2022deltacnn}. Compared to SBNet~\cite{ren2018sbnet}, our mask is directly derived from the difference between the original image and the edited image. Additionally, our method does not require any auxiliary network or extra model training. We also introduce other optimizations, such as normalization removal, kernel fusions and attention query pruning, to better adapt our engine for image editing.

\section{Method}
\lblsect{Method}

\looseness=-1 We build our method based on the following observation: during interactive image editing, a user often only changes the image content gradually. As a result, only a small subset of pixels in a local region is being updated at any moment. Therefore, we can reuse the activations of the original image for the unedited regions. As shown in \fig{method}, we first pre-compute all activations of the original input image. During the editing process, we locate the edited regions by computing a difference mask between the original and edited image. We then reuse the pre-computed activations for the unedited areas and only update the edited regions by applying convolutional filters to them. In \sect{Activation Sparsity}, we show the sparsity in the intermediate activations and present our main algorithm. In \sect{Sparse Engine}, we discuss the technical details of how our \engine (\engineabbr) supports the sparse inference and converts the theoretical computation reduction to measured speedup on hardware.

\subsection{Activation Sparsity}
\lblsect{Activation Sparsity}

\looseness=-1
\myparagraph{Preliminary.}
First, we closely study the computation within a single layer. We denote $\actori_l$ and $\actedi_l$ as the input tensor of the original image and edited image to the $l$-th convolution layer $\conv_l$, respectively. $\weight_l$ and $\bias_l$ are the weight and bias of $\conv_l$. The output of $\conv_l$ with input $\actedi_l$ could be computed in the following way due to the linearity of convolution:
\begin{align*}
    \conv_l(\actedi_l) &= \weight_l * \actedi_l+\bias_l \\
    &= \weight_l * (\actedi_l-\actori_l)+(\weight_l * \actori_{l}+\bias_l) \\
    & = \weight_l * \underbrace{\actdel_l}_\text{sparse}+\underbrace{\conv_l(\actori_l)}_\text{pre-computed},
\end{align*}
where $*$ is the convolution operator and $\actdel_l = \actedi_l-\actori_l$. If we have already pre-computed all the $\conv_l(\actori_l)$, we only need to compute $\weight_l*\actdel_l$. {\Naive}ly, computing $\weight_l * \actdel_l$ has the same complexity as $\weight_l * \actedi_l$. However, since the edited image shares similar features with the original image given a small edit, $\actdel_l$ should be sparse. Below, we discuss different strategies to leverage the activation sparsity to accelerate model inference. 

Our first attempt was to prune $\actdel_l$ by zeroing out elements smaller than a certain threshold to achieve the target sparsity. Unfortunately, this pruning method fails to achieve measured speedup due to the overheads of the on-the-fly pruning and irregular sparsity pattern. 

\looseness=-1
\myparagraph{Structured sparsity.} 
Fortunately, user edits are often highly structured and localized. As a result, $\actdel_l$ should also share the \emph{structured spatial sparsity}, where non-zero values are mostly aggregated within the edited regions, as shown in \fig{activation-difference}. We then directly use the original and edited images to compute a difference mask and sparsify $\actdel_l$ with this mask. 

\begin{figure*}[t]
    \centering
    \includegraphics[width=\linewidth]{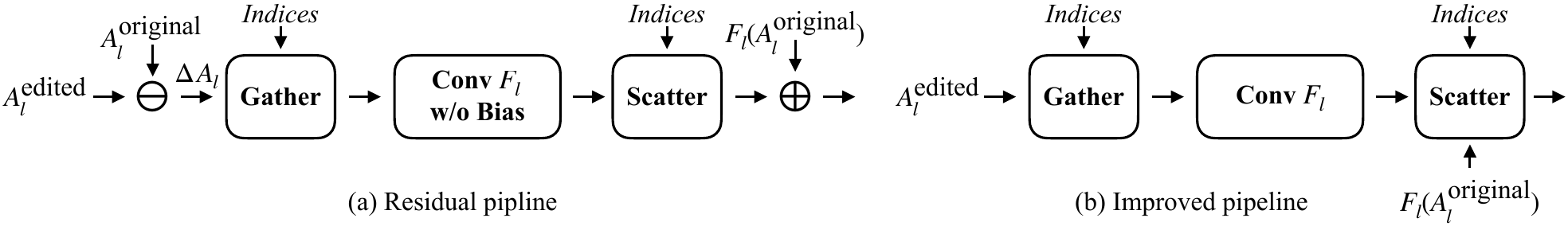}
    \vspace{-15pt}
    \caption{
    	\looseness=-1
    	Titling-based sparse convolution pipelines. (a) We first compute the activation difference $\actdel_l$ and gather the active blocks along the batch dimension from it according to the indices reduced from the difference mask. We then feed the blocks into the convolution $\conv_l$ without bias, scatter the output into a zero tensor, and add the residual $\conv_l(\actori_l)$ back. (b) We directly gather the blocks from $\actedi_l$ without computing $\actdel_l$. $\conv_l$ is computed with bias. We scatter the output into $\conv_l(\actori_l)$ instead of a zero tensor.
    }
    \vspace{-10pt}
    \lblfig{pipeline}
\end{figure*}

\begin{figure*}[t]
    \centering
    \includegraphics[width=\linewidth]{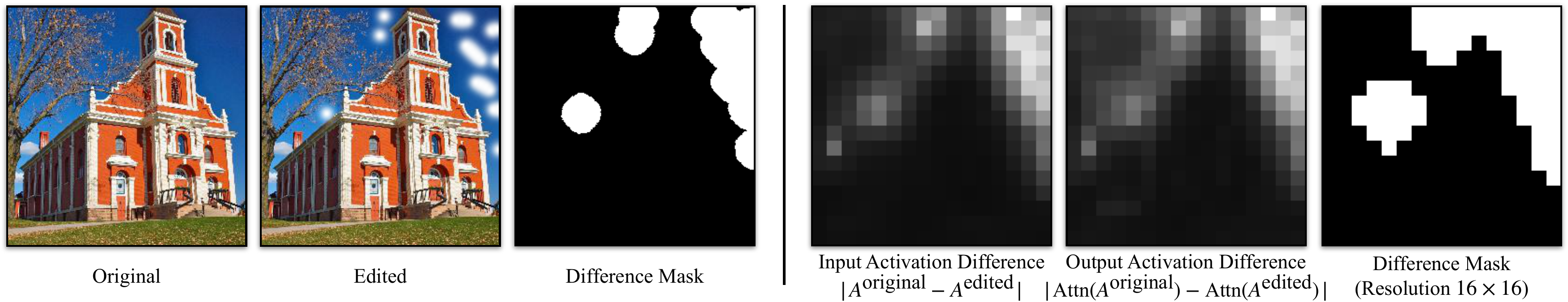}
    \vspace{-15pt}
    \caption{
    \looseness=-1
    The input and output activation differences of a $16\times 16$ self-attention layer in the DDPM model. Left: Detailed edit example with the difference mask. Right: Activation differences with the downsampled difference mask. \textit{Attn} is the self-attention layer. Brighter colors refer to larger differences. Both the input and output differences match the mask well.}
    \vspace{-15pt}
    \lblfig{attention}
\end{figure*}

\subsection{Sparse Engine \engineabbr}
\lblsect{Sparse Engine}
But how could we leverage the structured sparsity to accelerate $\weight_l*\actdel_l$? A \naive approach is to crop a rectangular edited region out of $\actdel_l$ for each convolution and only compute features for the cropped regions. Unfortunately, this \naive cropping method works poorly for the irregular edited regions (\eg, the example shown in \fig{activation-difference}). 

\looseness=-1 
\myparagraph{Tiling-based sparse convolution.}
Instead, as shown in \fig{pipeline}(a), we use a tiling-based sparse convolution algorithm. We first downsample the difference mask to different scales and dilate the downsampled masks (width 1 for diffusion models and 2 for GauGAN). Then we divide $\actdel_l$ into multiple small blocks of the same size spatially and index the difference mask at the corresponding resolution. Each block index refers to a single block with non-zero elements. We then gather the non-zero blocks (i.e., \emph{active blocks}) along the batch dimension and feed them into the convolution $\conv_l$. Finally, we scatter the output blocks into a zero tensor according to the indices to recover the original spatial size and add the pre-computed residual $\conv_l(\actori_l)$ back. The gathered active blocks overlap with width 2 for $3\times 3$ convolution with stride 1 to ensure the output blocks of the adjacent input blocks are seamlessly stitched together~\cite{ren2018sbnet}.

This pipeline in \fig{pipeline}(a) is equivalent to a simpler pipeline in \fig{pipeline}(b). Instead of gathering $\actdel_l$, we could directly gather $\actedi_l$. The convolution needs to be computed with bias $\bias_l$. Besides, we need to scatter the output blocks into $\conv_l (\actori_l)$ instead of a zero tensor. Thus, we do not need to store $\actori_l$ anymore, which further saves memory and removes the overheads of addition and subtraction. \fig{method} visualizes the pipeline.

However, the aforementioned pipeline still fails to produce a noticeable speedup due to extra kernel calls and memory movement overheads in {\ttfamily Gather} and {\ttfamily Scatter}. For example, the original dense $3\times3$ convolution with 128 channels and input resolution $256 \times 256$ takes 0.78ms on NVIDIA RTX 3090. The sparse convolution using pipeline \fig{pipeline}(b) on the example shown in \fig{activation-difference} (15.5\% edited regions) still needs 0.42ms in total, with the {\ttfamily Gather} and {\ttfamily Scatter} operations accounting for a significant overhead of 0.17ms (41\%). To mitigate these overheads, we further optimize \engineabbr by pre-computing normalization parameters and applying kernel fusion. Additionally, we extend \engineabbr to support attention layers.

\looseness=-1
\myparagraph{Pre-computing normalization parameters.}
For batch normalization~\cite{ioffe2015batch}, it is easy to remove the normalization layer during inference time since we can use pre-computed mean and variance statistics from model training. However, recent generative models often use instance normalization~\cite{ulyanov2016instance,huang2017arbitrary} or group normalization~\cite{wu2018group,nichol2021improved}, which compute the statistics on the fly during inference. These normalization layers incur overheads as we need to estimate the statistics from the full-size tensors. However, as the original and edited images are quite similar given a small user edit, we assume $\actori_l \approx \actedi_l$. This allows us to reuse the statistics of $\actori_l$ for the normalization instead of recomputing them for $\actedi_l$. Thus, normalization layers could be replaced by simple {\ttfamily Scale+Shift} operations with pre-computed $\actori_l$ statistics.

\myparagraph{Kernel fusion.}
As mentioned before, both the {\ttfamily Gather} and {\ttfamily Scatter} operations introduce significant data movement overheads. To reduce it, we fuse several element-wise operations ({\ttfamily Scale+Shift} and {\ttfamily Nonlinearity}) into {\ttfamily Gather} and {\ttfamily Scatter}~\cite{ren2018sbnet,ding2021ios,jia2019taso} and only apply these element-wise operations to the active blocks (\ie, edited regions). Furthermore, we perform the in-place computation to reduce the number of kernel calls and memory allocation overheads. 

In {\ttfamily Scatter}, we need to copy the pre-computed activation $\conv_l(\actori_l)$. This copying operation is highly redundant, as most elements from $\conv_l(\actori_l)$ do not involve any computation given a small edit and will be discarded in the next {\ttfamily Gather}. To reduce the tensor copying overheads, we fuse the {\ttfamily Scatter} with the following {\ttfamily Gather} by directly gathering the active blocks from $\conv_l(\actori_l)$ and the input blocks to be scattered. Sometimes, the residual connection in the ResBlock~\cite{he2016deep} contains a shortcut $1\times1$ convolution to match the channel number of the residual and the ResBlock output. We also fuse the {\ttfamily Scatter} in the shortcut branch, main branch, and the residual addition together to avoid the tensor copying overheads in the shortcut {\ttfamily Scatter}. Please refer to \app{Kernel Fusion} for more details.

\begin{figure*}[t]
    \centering
    \includegraphics[width=0.9\linewidth]{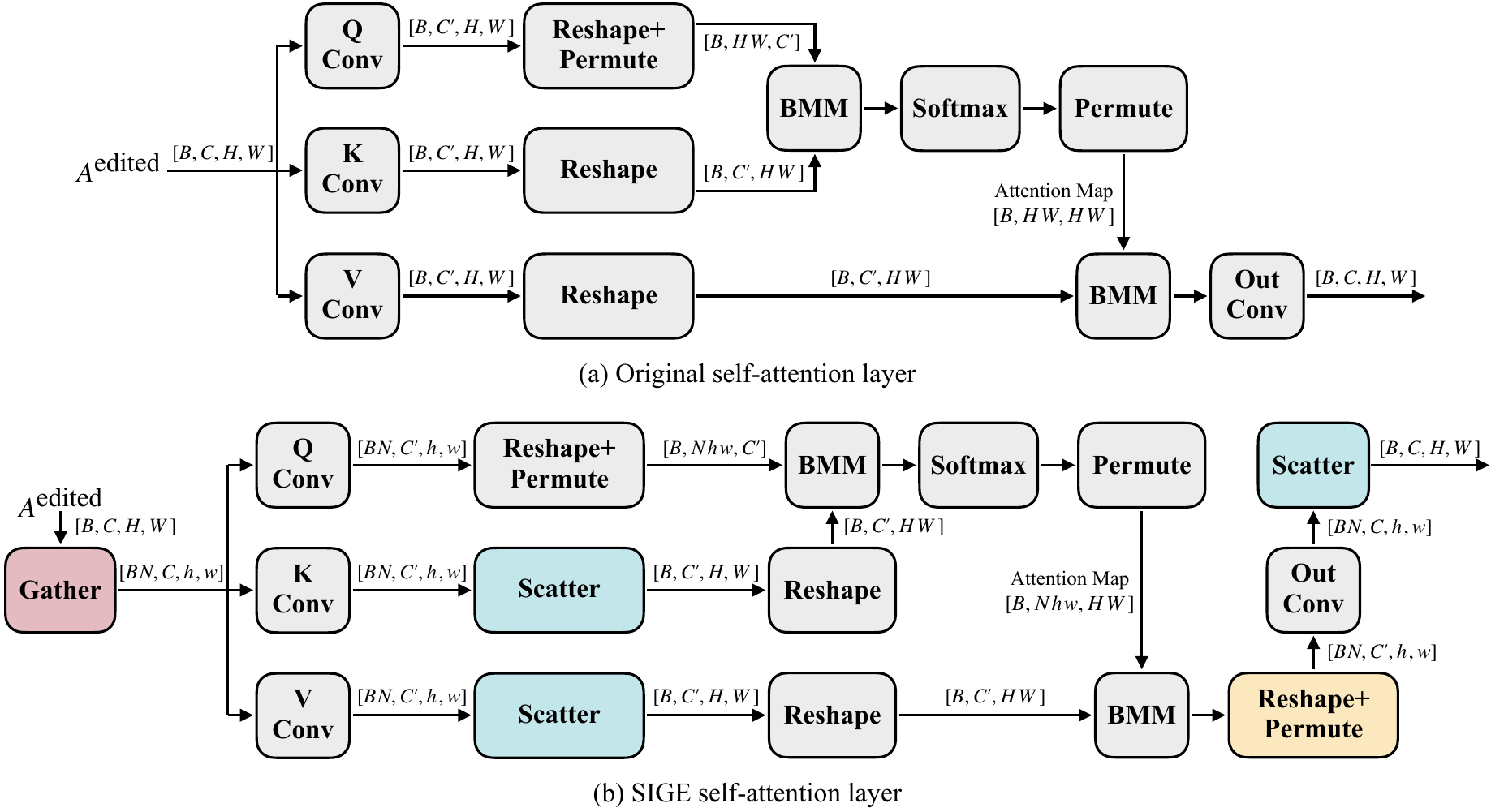}
    \vspace{-5pt}
    \caption{
    	\looseness=-1
    	The original self-attention layer \vs SIGE self-attention layer. All convolutions are $1\times1$, with batch size $B$, and the number of input and hidden channels $C$ and $C'$, respectively. The height and width of the input are represented by $H$ and $W$. $N$ denotes the number of active blocks, and $h$ and $w$ denote the height and width of the active blocks. For simplicity, we omit the original activation and active indices input to \texttt{Gather} and \texttt{Scatter}. The original pipeline computes the attention map for all query tokens. Our SIGE pipeline instead first gathers the active blocks and feeds them to the Q, K, and V convolutions. Then, only the keys and values are scattered back. Thus, only the edited query tokens are preserved, so the attention map size is reduced. After the Out Conv, the result tensor is scattered back into the original activation to obtain the final output.
    }
    \vspace{-15pt}
    \lblfig{sige-attention}
\end{figure*}

\myparagraph{Extension to attention layers.} \lblsect{Extension to attention layers.}
Recent models have adopted attention layers to enhance the image quality~\cite{zhang2019self,vaswani2017attention} and controllability~\cite{rombach2022high}. These attention layers could model long-range dependencies across image regions, potentially introducing non-local changes. 
However, when the edited region is small, it only has limited impacts on the unedited areas. \fig{attention} illustrates the difference maps for both the input and output activations of a $16\times16$ self-attention layer in the DDPM model, both of which closely match the difference mask. This indicates that user edits often change activations locally, even with attention layers, and our method's assumption still holds.

\looseness=-1
Based on this observation, we extend \engineabbr to attention layers. \fig{sige-attention} illustrates the difference between the vanilla and \engineabbr self-attention layer. As the unedited regions remain mostly unchanged, there is no need to compute the attention map for these areas. In our pipeline, \texttt{Gather} selectively prunes the unedited regions, only preserving the edited ones. After the Q, K, and V convolutions, we only scatter the keys and values back into the original activations. Thus, the query token number is reduced according to the edit ratio, and more importantly, it also reduces the size of the attention map correspondingly. Such memory reduction can lead to almost linear speedup even on an NVIDIA RTX 3090 GPU, as shown in \fig{teaser} and \fig{stable-diffusion}.

\section{Experiments}
\lblsect{Experiments}

\renewcommand \arraystretch{1.}
\begin{table*}[t]
\setlength{\tabcolsep}{5pt}
\scriptsize \centering
\begin{tabular}{llcccccccc}
\toprule
\multirow{2}{*}[\multirowcenter]{Model} & \multirow{2}{*}[\multirowcenter]{Method} &\multicolumn{2}{c}{MACs} & \multicolumn{2}{c}{PSNR ($\uparrow$)} & \multicolumn{2}{c}{LPIPS ($\downarrow$)} & \multirow{2}{*}[\multirowcenter]{FID ($\downarrow$)} & \multirow{2}{*}[\multirowcenter]{mIoU ($\uparrow$)} \\
\cmidrule(lr){3-4} \cmidrule(lr){5-6} \cmidrule(lr){7-8}
& & Value & Ratio & with G.T. & with Orig. & with G.T. & with Orig. \\
\midrule
\multirow{5}{*}[\multirowcenter]{DDPM} & Original & 249G & -- & 26.8 & -- & 0.069 & --& 65.4 & --\\
\cmidrule{2-10}
& 40\% Pruning & -- & -- & 24.9 & 31.0 & 0.991 & 0.101 & 72.2 & -- \\
\cmidrule{2-10}
& Patch & 72.0G & 3.5$\times$ & 26.8 & 40.6 & 0.076 & 0.022 & 66.4 & -- \\
\cmidrule{2-10}
& \textbf{Ours} & \textbf{65.3G} & \textbf{3.8}$\times$ & \textbf{26.8} & \textbf{52.4} & \textbf{0.070} & \textbf{0.009} & \textbf{65.8} & -- \\
\midrule
& Original & 66.9G & -- & 21.9 & -- & 0.143 & -- & 90.0 & --\\
\cmidrule{2-10}
PD & 40\% Pruning & -- & -- & 21.6 & 37.6 & 0.164 & 0.051 & 101 & --  \\
\cmidrule{2-10}
& \textbf{Ours} & \textbf{32.5G} & \textbf{2.1}$\times$ & \textbf{21.9} & \textbf{60.7} & \textbf{0.154} & \textbf{0.003} & \textbf{90.1} & -- \\
\midrule
\multirow{9}{*}{GauGAN} & Original & 281G & -- & 15.8 & -- & 0.409 & -- & 55.4 & 62.4 \\
\cmidrule{2-10}
& GAN Comp.~\cite{li2020gan} & 31.2G & 9.0$\times$ & 15.8 & 19.5 & \textbf{0.412} & 0.288 & 55.5 & 61.5 \\
\cmidrule{2-10}
& \textbf{Ours} & \textbf{30.7G} & \textbf{9.2}$\times$ & \textbf{15.8} & \textbf{26.5} & 0.413 & \textbf{0.113} & \textbf{54.4} & \textbf{62.1} \\
\cmidrule[0.9pt]{2-10}
& 0.19 GauGAN & 13.3G & 21$\times$ & 15.5 & 18.6 & 0.424 & 0.322 & 57.9 & 53.5 \\
\cmidrule{2-10}
 & GAN Comp. (S) & 9.64G & 29$\times$ & 15.7 & 19.1 & 0.422 & 0.310 & \textbf{50.4} & 57.4 \\
\cmidrule{2-10}
& \textbf{GAN Comp.+Ours} & \textbf{7.06G} & \textbf{40}$\times$ & \textbf{15.7} & \textbf{19.2} & \textbf{0.416} & \textbf{0.299} & 54.6 & \textbf{60.0}  \\
\midrule
&Original & 805G & -- & 19.3 & -- & 0.153 & --& 27.2 & --\\
\cmidrule{2-10}
Stable Diffusion & 50\% Pruning & -- & -- & \textbf{20.5} & \textbf{20.7} & 0.172 & 0.149 & 36.6 & --\\
\cmidrule{2-10}
& \textbf{Ours} & \textbf{387G} & \textbf{2.1}$\times$ & 19.2 & 19.9 & \textbf{0.157} & \textbf{0.126} & \textbf{26.8} & --\\
\bottomrule
\end{tabular}
\vspace{-5pt}
\caption{
	\looseness=-1 
	Quantitative evaluation. MACs measures the average computation for a single model forward over the entire dataset. PSNR/LPIPS \textit{with G.T.} means computing the metrics with the ground-truth images, and \textit{with Orig.} means computing with the generated samples from the original model. \textit{$P\%$ Pruning: Uniformly pruning $P\%$ model weights without fine-tuning.} \textit{Patch}: Cropping the smallest image patch that covers all the edited regions and blending the output patch into the original image. \textit{0.19 GauGAN}: Uniformly reducing each layer of GauGAN to 19\% channels and training from scratch. \textit{GAN Comp. (S)}: GAN Compression with a larger compression ratio. For all models, our method outperforms other baselines with less computation.}
\vspace{-15pt}
\lbltab{quality}
\end{table*}

Below we first describe our experiment setups, including models, baselines, datasets, and evaluation protocols. We then discuss our main qualitative and quantitative results. Finally, we include a detailed ablation study regarding the importance of each algorithmic design. 

\subsection{Setups}

\looseness=-1
\myparagraph{Models.}
We conduct experiments on the following four models, including diffusion models and GAN-based models, to explore the generality of our method.
\begin{itemize}[leftmargin=*]
	\item \emph{DDPM}~\cite{ho2020denoising} is a diffusion probabilistic model that models the data distribution through an iterative denoising process. It adopts a U-Net~\cite{ronneberger2015u} backbone for the denoising network. For fast sampling, we use \emph{DDIM}~\cite{song2020denoising} sampler to reduce the number of denosing steps from 1000 to 100. 
    \item \emph{Progressive Distillation (\progdist)}~\cite{salimans2021progressive} adopts network distillation~\cite{hinton2015distilling} to progressively reduce the number of steps for diffusion models.
    \item \emph{Stable Diffusion (\stbldiff)}~\cite{rombach2022high} is a text-to-image latent diffusion model~\cite{rombach2022high}. It uses a VAEGAN-based autoencoder to compress the image to a compact latent and applies the diffusion model in the latent space. The model includes multiple cross-attention layers to support text conditioning.
    \item \emph{GauGAN}~\cite{park2019semantic} is a paired image-to-image translation model which learns to generate a high-fidelity image given a semantic label map.
\end{itemize}

\looseness=-1
\myparagraph{Baselines.}
We compare our methods against the following baselines:
\begin{itemize}[leftmargin=*]
    \item \textit{Patch}. We crop the smallest patch covering all the edited regions, feed it into the model, and blend the output patch into the original image. 
    \item \looseness=-1 \textit{Crop}. For each convolution $\conv_l$, we crop the smallest rectangular region that covers all masked elements of the activation $\actedi_l$, feed it into $\conv_l$, and scatter the output patch into $\conv_l(\actori_l)$.
    \item \looseness=-1 \textit{$P\%$ Pruning.} We uniformly prune $P\%$ model weights without further fine-tuning. This is fair as our method uses pre-trained weights without fine-tuning. Since the fine-grained pruning is unstructured, it requires special hardware to achieve measured speedup, so we do not report MACs for this baseline.
    \item \looseness=-1 \textit{0.19 GauGAN}. We reduce each convolution layer of GauGAN to $19\%$ channels ($21\times$ MACs reduction) and train it from scratch.
    \item \textit{GAN Compression}~\cite{li2020gan}. A general-purpose compression method for conditional GANs. \textit{GAN Comp. (S)} means GAN Compression with a larger compression ratio.
    \item \textit{0.5 Original} means linearly scaling each layer of the original model to 50\% channels, and we only use this to benchmark our efficiency results.
\end{itemize}

\looseness=-1
\myparagraph{Datasets.}
We use the following three datasets:
\begin{itemize}[leftmargin=*]
    \item \emph{LSUN Church}. 
    We use the LSUN Church Outdoor dataset~\cite{yu15lsun} and follow the same preprocessing steps as prior works~\cite{ho2020denoising,song2020score}.
    To automatically generate a stroke editing benchmark, we first use Detic~\cite{zhou2021detecting} to segment the images in the validation set. For each segmented object, we use its segmentation mask to inpaint the image by CoModGAN~\cite{zhao2021comodgan} and treat the inpainted image as the original image. We generate the corresponding user strokes by first blurring the masked regions with the median filter and quantizing it into 6 colors following SDEdit~\cite{meng2022sdedit}. We collect 454 editing pairs in total  (431 synthetic + 23 manual). We evaluate DDPM~\cite{ho2020denoising,song2020denoising} and \progdist~\cite{salimans2021progressive} on this dataset.
    \item \looseness=-1 \emph{Cityscapes}. The dataset~\cite{cordts2016cityscapes} contains images of German street scenes. The training and validation sets consist of 2,975 and 500 images, respectively. Our editing dataset has 1,505 editing pairs in total. We evaluate GauGAN~\cite{park2019semantic} on this dataset.
    \item \looseness=-1 \emph{LAION}. We select a random subset of 1,000 images from LAION-5B~\cite{schuhmann2022laion} that meet or exceed $1024\times1024$ resolution. These images are then center-cropped and resized to $512 \times 512$ resolution. Next, we randomly occlude between $1\%$ and $25\%$ of the image area using circular masks. We use Stable Diffusion~\cite{rombach2022high} to inpaint the region and evaluate the visual quality of the output.
    
\end{itemize}
Please refer to \app{Benchmark Datasets} for more details about the benchmark datasets. 

\myparagraph{Metrics.}
Following previous works~\cite{meng2022sdedit,li2020gan,park2019semantic}, we use the standard metrics Peak Signal Noise Ratio (PSNR, higher is better), LPIPS (lower is better)~\cite{zhang2018perceptual}, and \fid (FID, lower is better)~\cite{heusel2017gans,parmar2021cleanfid}\footnote{We use \href{https://github.com/GaParmar/clean-fid}{clean-fid} for FID calculation.} to evaluate the image quality. For Cityscapes, we additionally adopt a semantic segmentation metric to evaluate the generated images. Specifically, we run DRN-D-105~\cite{yu2017dilated} on the generated images and compute the mean Intersection over Union~(mIoU) of the segmentation results. Generally, a higher mIOU indicates that the generated images look more realistic and better align with the input.

\begin{figure*}[!t]
    \centering
    \includegraphics[width=0.93\linewidth]{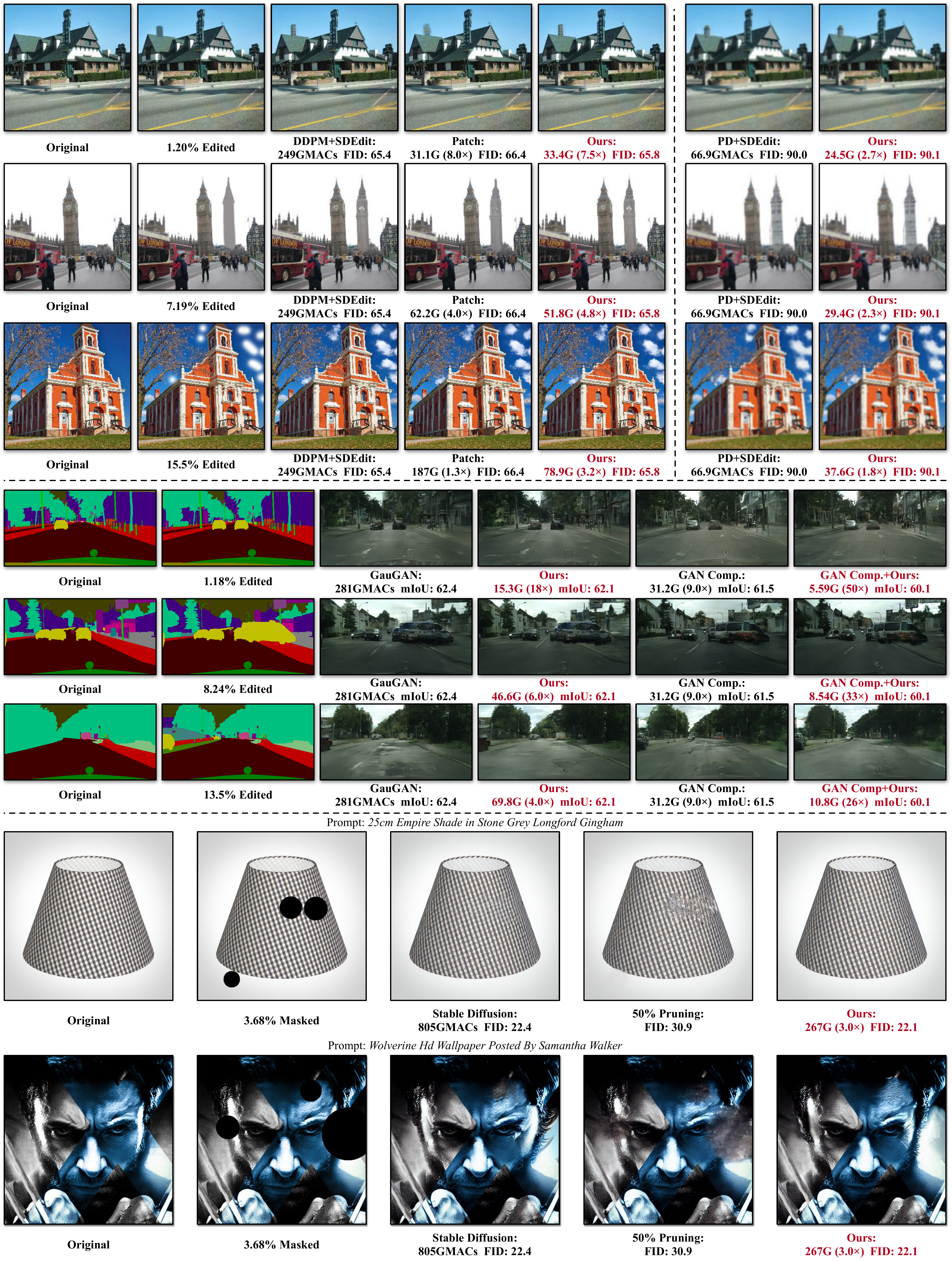}
    \vspace{-10pt}
    \caption{Qualitative results of our method under different edit or mask sizes. MACs measure the computation for a single model forward. Our method well preserves the visual fidelity of the original model without losing global context. On the contrary, \textit{Patch} (cropping the smallest image patch that covers all the edited regions and scattering the output patch back into the original image) performs poorly because of the lack of global context when the edit is small.}
    \vspace{-15pt}
    \lblfig{quality}
\end{figure*}

\looseness=-1 
\myparagraph{Implementation details.} 
We use DDIM sampler~\cite{song2020denoising} for both DDPM and Stable Diffusion (\stbldiff). Specifically, the number of total denoising steps for DDPM, \stbldiff, and Progressive Distillation (\progdist) are 100, 50, and 8, respectively, and we use 50, 40, and 5 steps for SDEdit~\cite{meng2022sdedit}. We dilate the difference mask by 5, 5, 2, 5, and 1 pixels for DDPM, \stbldiff, \progdist with resolution 128, \progdist with resolution 256, and GauGAN, respectively. For \stbldiff decoder, we dilate the difference mask by 45 pixels. Besides, we apply \engineabbr to all convolutional layers whose input feature map resolutions are larger than $32\times 32$,  $16\times 16$, $8 \times 16$ and $16 \times 32$ for DDPM,  \progdist, original GauGAN, and GAN Compression, respectively. For \stbldiff, we apply \engineabbr to all convolutional layers and attention layers except ones in the middle stages. As the attention layers in DDPM and \progdist only consume a small portion of the overall latency, we do not apply \engineabbr to them. For diffusion models, we pre-compute and reuse the statistics of the original image for all group normalization layers~\cite{wu2018group}. For GANs, we pre-compute and reuse the statistics of the original image for all instance normalization layers~\cite{ulyanov2016instance} whose resolution is higher than $16 \times 32$. For all models, the sparse block size for $3\times3$ convolution is 6, and $1\times1$ convolution is 4. All results are measured with FP32 precision.

\begin{figure*}[t]
    \centering
    \includegraphics[width=\linewidth]{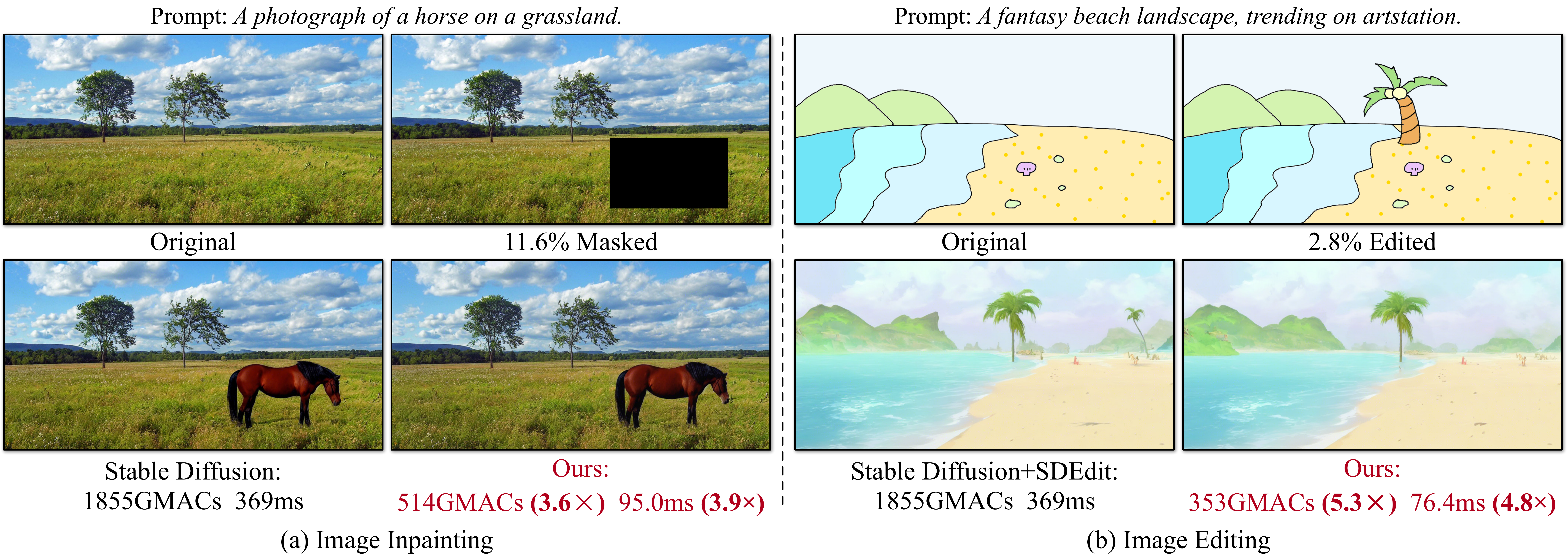}
    \caption{
    	Qualitative results of Stable Diffusion~\cite{rombach2022high}. The computation and latency are measured for a single diffusion step on NVIDIA RTX 3090. For image inpainting, with $11.6\%$ masked regions, our method reduces the computation by $3.6\times$, resulting in a $3.9\times$ speedup. For image editing, we reduce the computation by $5.3\times$, achieving a $4.8\times$ speedup with a $2.8\%$-area edit.
    }
    \vspace{-5pt}
    \lblfig{stable-diffusion}
\end{figure*}
\renewcommand \arraystretch{1.}
\begin{table*}[t]
\setlength{\tabcolsep}{4.5pt}
\scriptsize \centering
\begin{tabular}{lcccccccccccccc}
\toprule
\multirow{2}{*}[\multirowcenter]{Model} & \multirow{2}{*}[\multirowcenter]{Edit Size} & \multirow{2}{*}[\multirowcenter]{Method} & \multicolumn{2}{c}{MACs} & \multicolumn{2}{c}{3090} & \multicolumn{2}{c}{2080Ti} & \multicolumn{2}{c}{Intel Core i9} & \multicolumn{2}{c}{M1 Pro CPU} & \multicolumn{2}{c}{M1 Pro GPU} \\
\cmidrule(lr){4-5} \cmidrule(lr){6-7} \cmidrule(lr){8-9} \cmidrule(lr){10-11} \cmidrule(lr){12-13} \cmidrule(lr){14-15} 
& & & Value & Ratio & Value & Ratio & Value & Ratio & Value & Ratio & Value & Ratio & Value & Ratio \\

\midrule
\multirow{9}{*}[\multirowcenter]{DDPM} & \multirow{2}{*}[\multirowcenter]{--} & Original & 248G & -- & 37.5ms & -- & 54.6ms & -- & 609ms & -- & 12.9s & -- & 183ms & -- \\
\cmidrule{3-15}
& & 0.5 Original & 62.5G & 4.0$\times$ & 20.0ms & 1.9$\times$ & 31.2ms & 1.8$\times$ & 215ms & 2.8$\times$ & 3.22s & 4.0$\times$ & 90.5ms & 2.0$\times$ \\
\cmidrule{2-15}
& \multirow{2}{*}[\multirowcenter]{1.20\%} & Crop & \textbf{32.6G} & \textbf{7.6}$\times$ & 15.5ms & 2.4$\times$ & 29.3ms & 1.9$\times$ & 185ms & 3.3$\times$ & \textbf{1.85s} & \textbf{6.9}$\times$ & 52.1ms & 3.5$\times$\\
\cmidrule{3-15}
& & Ours & 33.4G & 7.5$\times$ & \textbf{12.6ms} & \textbf{3.0}$\times$ & \textbf{19.1ms} & \textbf{2.9}$\times$ & \textbf{147ms} & \textbf{4.1}$\times$ & 1.96s & 6.6$\times$ & \textbf{39.5ms} & \textbf{4.6}$\times$ \\
\cmidrule{2-15}
& \multirow{2}{*}[\multirowcenter]{15.5\%} & Crop & 155G & 1.6$\times$ & 30.5ms & 1.2$\times$ & 44.5ms & 1.2$\times$ & 441ms & 1.4$\times$ & 8.09s & 1.6$\times$ &144ms & 1.3$\times$ \\
\cmidrule{3-15}
& & Ours & 78.9G & 3.2$\times$ & 19.4ms & 1.9$\times$ & 29.8ms & 1.8$\times$ & 304ms & 2.0$\times$ & 5.04s & 2.6$\times$ & 75.8ms & 2.4$\times$ \\
\midrule

\multirow{5}{*}[\multirowcenter]{PD256} & \multirow{2}{*}[\multirowcenter]{--} & Original & 119G & -- & 35.1ms & -- & 51.2ms & -- & 388ms & -- & 6.18s & --& 178ms & --  \\
\cmidrule{3-15} 
& & 0.5 Original & 31.0G & 3.8$\times$ & 29.4ms & 1.2$\times$ & 43.2ms & 1.2$\times$ & 186ms & 2.1$\times$ & 1.72s & 3.6$\times$ & 151ms & 1.2$\times$  \\
\cmidrule{2-15}
& 1.20\% & Ours & \textbf{25.9G} & \textbf{4.6}$\times$ & \textbf{18.6ms} & \textbf{1.9}$\times$ & \textbf{26.4ms} & \textbf{1.9}$\times$ & \textbf{152ms} & \textbf{2.5}$\times$ & \textbf{1.55s} & \textbf{4.0}$\times$ & \textbf{59.9ms} & \textbf{3.0}$\times$   \\
\cmidrule{2-15}
& 15.5\% & Ours & 48.5G & 2.5$\times$ & 21.4ms & 1.6$\times$ & 30.7ms & 1.7$\times$ & 250ms & 1.6$\times$ & 3.22s & 1.9$\times$ & 73.3ms & 2.4$\times$ \\
\midrule

\multirow{9}{*}[\multirowcenter]{GauGAN} & \multirow{2}{*}[\multirowcenter]{--} & Original & 281G & -- & 45.4ms & -- & 49.5ms & -- & 682ms & -- & 14.1s & -- & 151ms & -- \\
\cmidrule{3-15}
& & GAN Compression & 31.2G & 9.0$\times$ & 15.1ms & 3.0$\times$ & 25.0ms & 2.0$\times$ & 333ms & 2.1$\times$ & 2.11s & 6.7$\times$ & 75.3ms & 2.0$\times$ \\
\cmidrule{2-15}
& \multirow{2}{*}[\multirowcenter]{1.18\%} & Ours & 15.3G & 18$\times$ & \textbf{8.11ms} & \textbf{5.6}$\times$ & 19.3ms & 2.6$\times$ & 114ms & 6.0$\times$ & 0.990s & 14$\times$ & 29.1ms & 5.2$\times$ \\
\cmidrule{3-15}
& & GAN Comp.+Ours & \textbf{5.59G} & \textbf{50}$\times$ & 8.72ms & 5.2$\times$ & \textbf{16.2ms} & \textbf{3.1}$\times$ & \textbf{53.1ms} & \textbf{13}$\times$ & \textbf{0.370s} & \textbf{38}$\times$ & \textbf{25.6ms} & \textbf{5.9}$\times$ \\
\cmidrule{2-15}
& \multirow{2}{*}[\multirowcenter]{13.5\%} & Ours & 69.8G & 4.0$\times$ & 17.8ms & 2.5$\times$ & 27.1ms & 1.8$\times$ & 238ms & 2.9$\times$ & 4.06s & 3.5$\times$ & 89.1ms & 1.7$\times$\\
\cmidrule{3-15}
& & GAN Comp.+Ours & 10.8G & 26$\times$ & 10.0ms & 4.5$\times$ & 17.4ms & 2.8$\times$ & 94.4ms & 7.2$\times$ & 0.741s & 19$\times$ & 45.5ms & 3.3$\times$ \\
\midrule

& \multirow{2}{*}[\multirowcenter]{--} & Original & 1855G & -- & 369ms & -- & -- & -- & 8.93s & -- & -- & -- & -- & -- \\
\cmidrule{3-15}
Stable& & 0.5 Original & 593G & 3.1$\times$ & 279ms & 1.3$\times$ & -- & -- & 6.79s & 1.3$\times$ & -- & -- & -- & -- \\
\cmidrule{2-15}
Diffusion& 2.78$\%$ & Ours & \textbf{353G} & \textbf{5.3}$\times$ & \textbf{76.4ms} & \textbf{4.8}$\times$ & -- & -- & \textbf{1.37s} & \textbf{6.5}$\times$ & -- & -- & -- & -- \\
\cmidrule{2-15}
& 11.5$\%$ & Ours & 514G & 3.6$\times$ & 95.0ms & 3.9$\times$ & -- & -- & 2.35s & 3.5$\times$ & -- & -- & -- & -- \\
\bottomrule
\end{tabular}
\vspace{-5pt}
\caption{
    \looseness=-1 Measured computation and latency for a single model forward on different devices. The detailed edit examples are shown in \fig{quality} and \ref{fig:stable-diffusion}. \textit{0.5 Original}: Linearly scaling each layer of the model to 50\% channels. \textit{Crop}: For each convolution, we find the smallest patch covering the masked elements, crop it out, feed it into the convolution and scatter the output patch into the original image activation. Our method could reduce up to 18$\times$ MACs and achieve up to 5.6$\times$, $2.9\times$, 6.0$\times$, 14$\times$, and $5.2\times$ latency reductions on NVIDIA RTX 3090, 2080Ti, Intel Core i9-10920X and M1 Pro CPU and GPU. With GAN Compression, we could further speed up GauGAN by $9.5\times$ on Intel Core-i9 and $38\times$ on Apple M1 Pro CPU.
}
\vspace{-10pt}
\lbltab{efficiency}
\end{table*}

\subsection{Main Results}
\lblsect{Main Results}

\looseness=-1
\myparagraph{Image quality.}
We report the quantitative results of applying our method to DDPM~\cite{ho2020denoising,song2020denoising}, \progdist~\cite{salimans2021progressive}, GauGAN~\cite{park2019semantic}, and Stable Diffusion~\cite{rombach2022high} on SDEdit~\cite{meng2022sdedit} image editing and text-guided inpainting in \tab{quality}. \fig{quality} shows some qualitative results. For PSNR and LPIPS, \textit{with G.T.} means computing the metric with the ground-truth images. \textit{With Orig.} means computing the metric with the samples generated by the original model. On LSUN Church, we only use 431 synthetic images for the \textit{PSNR/LPIPS with G.T.} metrics, as manual edits do not have ground truths. For the other metrics, we use the entire LSUN Chur ch dataset (431 synthetic + 23 manual edits). On Cityscapes, we view the synthetic semantic maps as the original input and the ground-truth semantic maps as the edited input for the \textit{PSNR/LPIPS with G.T.} metrics, which has 1505 samples. For the other metrics, we include the symmetric edits (view the ground-truth semantic maps as the original inputs and synthetic semantic maps as the edited inputs), with 3010 samples in total. For the models with method \textit{Patch} and \textit{Ours}, whose computation is edit-dependent, we measure the average MACs over the whole dataset.

For SDEdit with DDPM and \progdist, our method outperforms all baselines consistently and achieves results on par with the original model. The \textit{Patch} inference fails when the edited region is small as the global context is insufficient. Although our method only applies convolutional filters to the local edited regions, it can reuse the global context stored in the original activations. Therefore, it performs the same as the original model. For GauGAN, our method also performs better than GAN Compression~\cite{li2020gan} with an even larger MACs reduction. When applying it to GAN Compression, we further achieve a $\sim 40\times$ MACs reduction with minor performance degradation, beating both \textit{0.19 GauGAN} and \textit{GAN Comp. (S)}. For 
Stable Diffusion, \engineabbr reduces its computation by $2.1\times$ on average while maintaining LPIPS~\cite{zhang2018perceptual} and FID~\cite{heusel2017gans,parmar2021cleanfid}. As illustrated in \fig{quality}, although \textit{50\% Pruning} achieves higher PSNR values, it significantly lags in visual quality compared to both the original model and our method. In addition to the automatically generated inpainting examples, we manually curate two $512\times 1024$ examples on image inpainting and image-to-image translation in \fig{stable-diffusion}. Our method closely mirrors the original model's results while reducing the cost by $4 \sim 5 \times$.

\looseness=-1 
\myparagraph{Model efficiency.}
For real-world interactive image editing applications, inference acceleration on hardware is more critical than computation reduction. To verify the effectiveness of our proposed engine, we measure the speedup of the edit examples  shown in \fig{quality} for DDPM, PD and GauGAN and \ref{fig:stable-diffusion} for Stable Diffusion on five devices, including NVIDIA RTX 3090, NVIDIA RTX 2080Ti, Intel Core i9-10920X CPU, and Apple M1 Pro CPU and GPU, with different computational powers. We use batch size 1 to simulate real-world use. For GPU devices, we first perform 200 warm-up runs and measure the average latency of the next 200 runs. For CPU devices, we perform 10 warm-up runs and 10 test runs, repeat this process 5 times and report the average latency. The results are shown in \tab{efficiency}. 

The original Progressive Distillation~\cite{salimans2021progressive} can only generate $128\times 128$ images, which is too small for real use. We add some extra layers to adapt the model to resolution $256 \times 256$. For the \textit{Crop} baseline, we also pre-compute the normalization parameters for fair comparisons. When the edit pattern is like a rectangle, this baseline reduces similar computation with ours (\eg, the first example of DDPM in \fig{quality}). However, the speedup is still worse than ours on various devices due to the large memory index overheads in native PyTorch. When the edited region is far from a rectangle (\eg, the third example of DDPM), the cropped patch has much redundancy. Therefore, even though only 15.5\% areas are edited, the MACs reduction is only 1.6$\times$. For Stable Diffusion, as mentioned in \sect{Extension to attention layers.}, by selectively pruning unedited queries, \engineabbr reduces the attention map size and memory overheads in attention layers. For $512 \times 1024$ image editing example in \fig{stable-diffusion}, the first self-attention layer of the diffusion network takes $64 \times 128$ input, resulting in an $8192 \times 8192$ attention map. \engineabbr reduces the attention map size to $672 \times 8192$ ($12\times$ smaller). Therefore, it can achieve a similar speedup ratio to the computation reduction even on GPUs.
\engineabbr achieves up to 5.6$\times$, 2.9$\times$, 6.0$\times$, 14$\times$, and 5.2$\times$ speedups on RTX 3090, 2080Ti, Intel Core i9-10920X, Apple M1 Pro CPU and GPU, respectively. When applied to GAN Compression, \engineabbr achieves 9.5$\times$ and 38$\times$ latency reductions on Intel Core i9 and Apple M1 Pro CPU, respectively. Additionally, we apply \engineabbr to the encoder and decoder part of Stable Diffusion. For the image editing example in \fig{stable-diffusion}, this results in an $8\times$ speedup for the encoder and a $5\times$ speedup for the decoder.

\looseness=-1 

\looseness=-1
\myparagraph{Large edits.} In \fig{curves}, we further analyze the DDPM's and GauGAN's computation/latency \vs edit ratio curves on both NVIDIA RTX 3090 and Intel Core i9. On both devices, computation and latency scale linearly with the edit ratio. For the RTX 3090, our method delivers speed improvements up to edit ratios between $65\%$ and $75\%$. On the Intel Core i9, the ceiling is approximately $45\%$. \fig{xl_edits} showcases examples with large edit ratios exceeding $50\%$, where \engineabbr still matches the original model in visual quality. Moreover, in real-world applications, users can break down large edits into smaller increments. Our method facilitates fast updates as these incremental edits are applied, as discussed below.

\myparagraph{Sequential edits.} In \fig{sequential}, we show the results of sequential edits with our method. Specifically, \textit{One-time Pre-computation} performs as well as the \textit{Full Model}, demonstrating that our method can be applied to multiple sequential edits with only one-time pre-computation in most cases. Moreover, for extremely large edits, we could use \engineabbr to incrementally update the pre-computed features (\textit{Incremental Pre-computation}) and condition the later edits on the recomputed one. Its results are also as good as the full model.

\subsection{Ablation Study}
\lblsect{Ablation Study}
Below we perform several ablation studies to show the effectiveness of each design choice.

\looseness=-1
\myparagraph{Memory usage.}
The pre-computed activations of the original image require additional memory storage. We profile the peak memory usage of the original model and our method in PyTorch. Our method only increases the peak memory usage of a single forward for DDPM~\cite{ho2020denoising}, \progdist~\cite{salimans2021progressive}, GauGAN~\cite{park2019semantic}, and GAN Compression~\cite{li2020gan} by 0.1G, 0.1G, 0.8G, and 0.3G, respectively with FP32 precision. Specifically, it needs to store additional 169M, 56M, 239M, 275M, and 120M parameters for DDPM, \progdist, Stable Diffusion~\cite{rombach2022high}, original GauGAN and GAN Compression, respectively, for a single forward. For the diffusion models, we need to store activations for all iteration steps (\eg, 50 for DDIM sampler~\cite{song2020denoising} and 5 for \progdist). However, data movement and kernel computation are asynchronous on GPU, so we could store the activations in CPU memory and load the on-demand ones on GPU to reduce peak memory usage.

\looseness=-1

\begin{figure}[t]
    \centering
    \includegraphics[width=\linewidth]{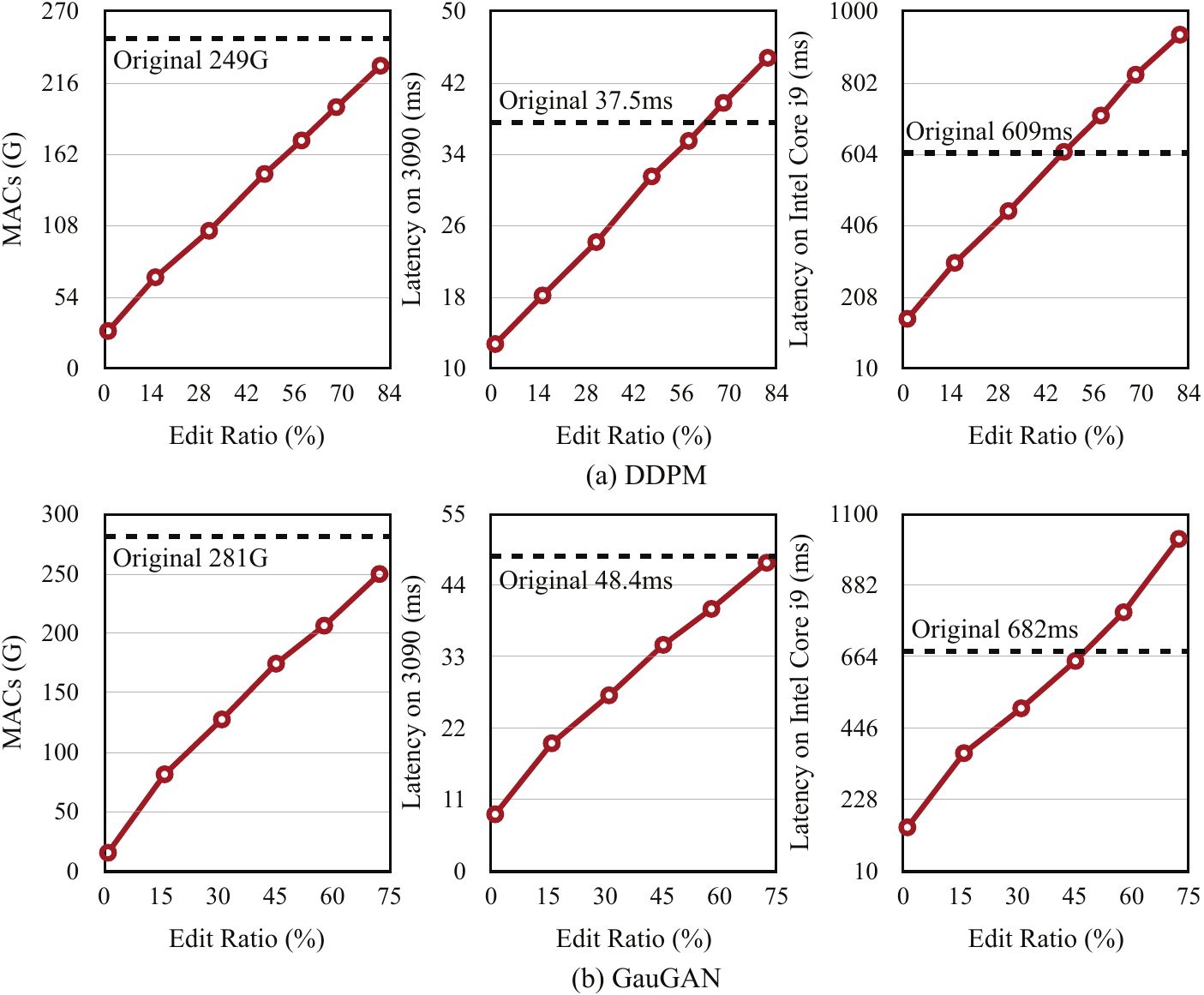}
    \vspace{-15pt}
    \caption{Computation/latency \vs edit ratio curves of \engineabbr for DDPM and GauGAN. The black dashed line represents the performance of the original model. Both computation and latency scale linearly with the edit ratio. On NVIDIA RTX 3090, our method attains speedups for edits up to $75\%$, and on Intel Core i9, up to $45\%$.}
    \lblfig{curves}
    \vspace{-15pt}
\end{figure}
\begin{figure*}[t]
    \centering
    \includegraphics[width=\linewidth]{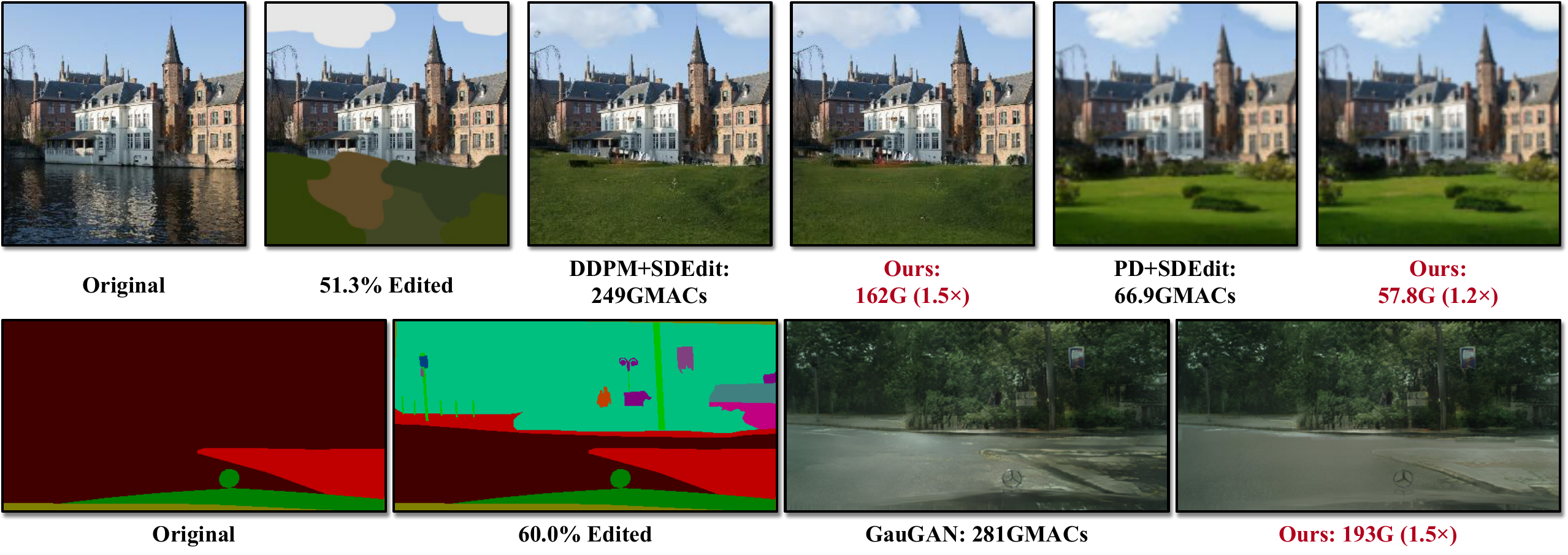}
    \vspace{-20pt}
    \caption{Qualitative results of \engineabbr with large edits. With $50\sim 60\%$ edits, \engineabbr can still preserve visual fidelity of the original model without losing global context while reducing the computation by up to $1.5\times$.}
    \lblfig{xl_edits}
\end{figure*}
\begin{figure}[t]
    \centering
    \includegraphics[width=\linewidth]{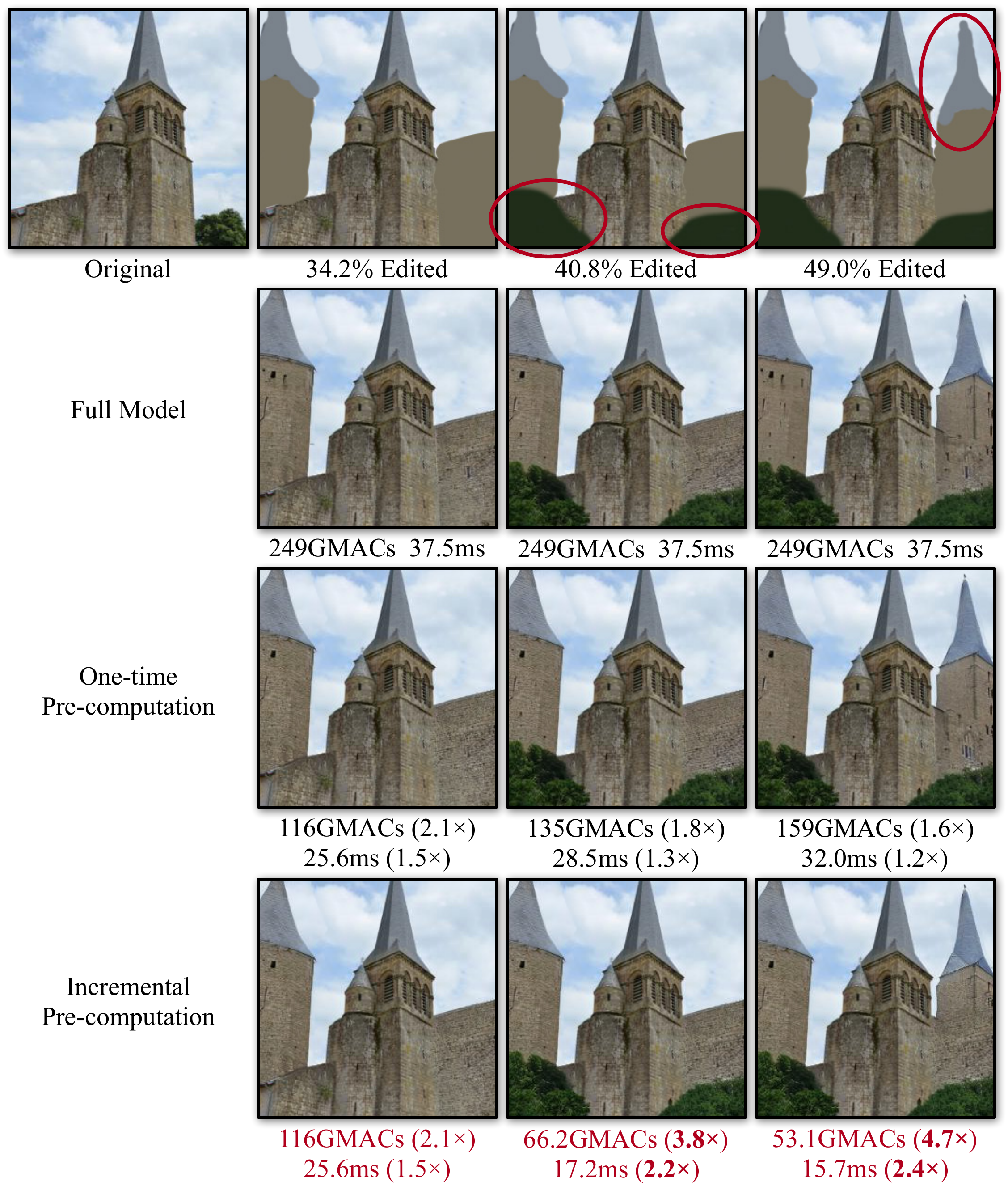}
    \vspace{-15pt}
    \caption{Sequential editing results with SIGE. The computation and latency are measured on NVIDIA RTX 3090 for a single forward. \textit{Full Model} means the results with the full model. \textit{One-time Pre-computation} means we pre-compute the original image features for all the edit steps. \textit{Incremental Pre-computation} means we incrementally update the pre-computed features with SIGE before the next edit step. The image quality of all methods is quite similar.}
    \lblfig{sequential}
    \vspace{-15pt}
\end{figure}
\begin{figure*}[t]
    \centering
    \includegraphics[width=\linewidth]{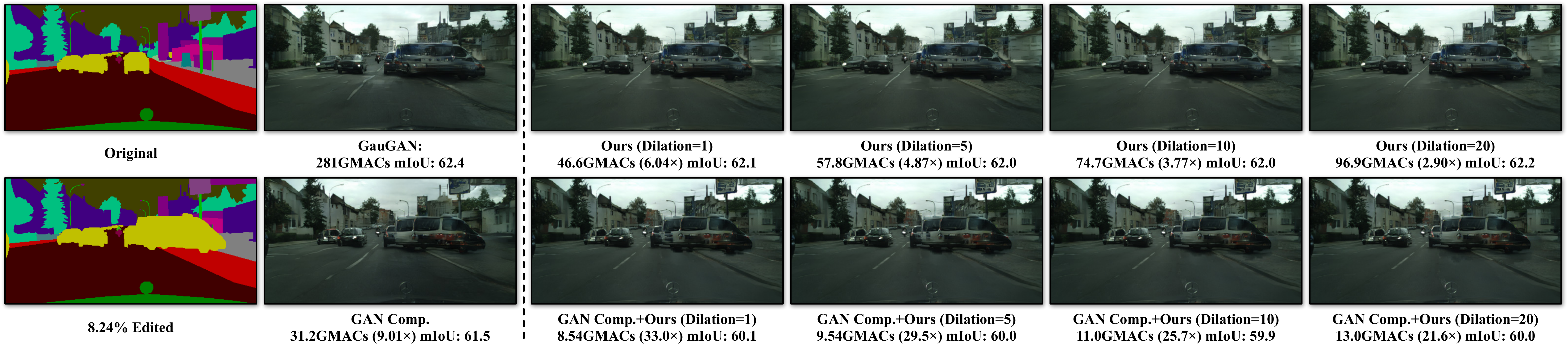}
    \vspace{-15pt}
    \caption{Visualization results of different dilation sizes on GauGAN. 
    Although without mIoU improvement, increasing the dilation could smoothly blend the boundary between the edited region and unedited regions to improve the image quality slightly. Specifically, the shadow boundary of the added car fades when dilation increases. However, it will incur more computations.}
    \lblfig{dilation}
    \vspace{-5pt}
\end{figure*}

\looseness=-1 
\myparagraph{Speedup of each design.}
\tab{ablation} shows the effectiveness of each optimization we add to \engineabbr. For DDPM on RTX 2080Ti, {\naive}ly applying the tiling-based sparse convolution can reduce the computation by 7.6$\times$. Still, the latency reduction is only 1.6$\times$ due to the large memory overheads in {\ttfamily Gather} and {\ttfamily Scatter}. Pre-computing the normalization parameters can remove the latency of normalization statistics calculation and reduce the overall latency to 29.6ms. Fusing element-wise operations into the {\ttfamily Gather} and {\ttfamily Scatter} can remove redundant operations in the unedited regions and also reduce the memory allocation overheads (about 9ms). Finally, fusing the {\ttfamily Scatter} and {\ttfamily Gather} to {\ttfamily Scatter-Gather} and {\ttfamily Scatter} in the shortcut branch and main branch can further reduce about 1.6ms tensor copying overheads, achieving a 2.9$\times$ speedup. For Stable Diffusion~\cite{rombach2022high} on RTX 3090, both the computation reduction ($1.6\times$) and the speedup  ($1.1\times$) are poor without our \engineabbr attention, as the large attention map incurs significant memory overheads. With our \engineabbr attention, we achieve a $5.3\times$ MACs reduction. Moreover, as we prune the unedited query tokens to reduce the attention map size accordingly, the memory overheads decrease correspondingly. Therefore, the speedup is much more prominent (4.8$\times$).

\looseness=-1 
\myparagraph{Experiments with TensorRT.} 
Real-world model deployment also depends on deep learning backends with optimized libraries and runtimes. To demonstrate the effectiveness and extensibility of \engineabbr, we also implement our kernels in a widely-used backend TensorRT\footnote{We benchmark the results with TensorRT 8.4.} and benchmark the DDPM latency results on RTX 2080Ti in \tab{backend}. Specifically, our speedup ratio becomes more prominent with TensorRT compared to PyTorch, especially for small edits, as TensorRT better supports small convolutional kernels with higher GPU utilization than PyTorch.

\myparagraph{Dilation hyper-parameter.} 
We show the results of our method with different dilation on GauGAN in \fig{dilation}. Increasing the dilation incurs more computations but also slightly improves the image quality. Specifically, the shadow boundary of the added car fades as the dilation increases. We choose dilation 1 since the image quality is almost the same as 20 while delivering the best speed.
\renewcommand \arraystretch{1.}
\begin{table}[t]
	\setlength{\tabcolsep}{3.5pt}
	\scriptsize \centering
	\begin{tabular}{lcccccccc}
		\toprule
		\multirow{2.5}{*}{Models}&\multirow{2.5}{*}{MACs}&  \multicolumn{5}{c}{Optimizations} & \multicolumn{2}{c}{Latency} \\
		\cmidrule{3-7} \cmidrule(lr){8-9}
		& & Sparse & Norm. & Elem. & Sct. & Attn. & Value & Ratio\\
		\midrule
		& 249G & & & & & & 54.6ms & -- \\
		\cmidrule{2-9}
		& & $\checkmark$ & & & & & 34.0ms & 1.6$\times$ \\
		\cmidrule{3-9}
		DDPM & 32.6G& $\checkmark$ & $\checkmark$ & & & & 29.6ms & 1.8$\times$ \\
		\cmidrule{3-9}
		& (7.6$\times$)& $\checkmark$ & $\checkmark$ & $\checkmark$ & & & 20.7ms & 2.6$\times$  \\
		\cmidrule{3-9}
		& & $\checkmark$ & $\checkmark$ & $\checkmark$ & $\checkmark$ & & 19.1ms & 2.9$\times$ \\
		\midrule
		& 1855G & & & & & & 369ms & -- \\
		\cmidrule{2-9}
		SD & 1193G (1.6$\times$) & $\checkmark$ & $\checkmark$ & $\checkmark$ & $\checkmark$ & & 335ms & 1.1$\times$ \\
		\cmidrule{2-9}
		& 353G (5.3$\times$) & $\checkmark$ & $\checkmark$ & $\checkmark$ & $\checkmark$ & $\checkmark$ & 76.4ms & 4.8$\times$ \\			
		\bottomrule
	\end{tabular}
	\vspace{-5pt}
	\caption{
		\looseness=-1 Ablation study of each optimization. \textbf{Sparse}: Using tiling-based sparse convolution. \textbf{Norm.}: Pre-computing normalization parameters. \textbf{Elem.}: Fusing element-wise operations. \textbf{Sct.}: Fusing {\ttfamily Scatter} to reduce the tensor copying overheads. \textbf{Attn.}: Using SIGE attention layers. With all optimizations, we could reduce the latency of DDPM by 2.9$\times$ on NVIDIA RTX 2080Ti and Stable Diffusion by 4.8$\times$ on RTX 3090.
	}
	\vspace{-5pt}
	\lbltab{ablation}
\end{table}
\renewcommand \arraystretch{1.}
\begin{table}[t]
	\setlength{\tabcolsep}{5.5pt}
	\scriptsize \centering
	\begin{tabular}{lccccccc}
		\toprule
		\multirow{2.5}{*}{Method} & \multirow{2.5}{*}{Edit Size} & \multicolumn{2}{c}{MACs} & \multicolumn{2}{c}{PyTorch} & \multicolumn{2}{c}{TensorRT} \\
		\cmidrule(lr){3-4} \cmidrule(lr){5-6} \cmidrule(lr){7-8}
		& & Value & Ratio & Value & Ratio & Value & Ratio \\
		\midrule
		Original & -- & 249G & -- & 54.6ms & -- & 47.7ms & -- \\
		\midrule
		& 1.20\% & 33.4G & 7.5$\times$ & 19.1ms & 2.9$\times$ & 14.4ms & \textbf{3.3}$\times$ \\
		\cmidrule{2-8}
		Ours & 7.19\% & 51.8G & 4.8$\times$ & 22.1ms & 2.5$\times$ & 18.6ms & \textbf{2.6}$\times$ \\
		\cmidrule{2-8}
		& 15.5\% & 78.9G & 3.2$\times$ & 29.8ms & \textbf{1.8}$\times$ & 26.9ms & \textbf{1.8}$\times$ \\
		\bottomrule
	\end{tabular}
	\vspace{-5pt}
	\caption{
		\looseness=-1 Latency comparisons of DDPM on RTX 2080Ti between PyTorch and TensorRT. The speedup ratio is larger in TensorRT than PyTorch, especially when the edit size is small.
	}
	\vspace{-10pt}
	\lbltab{backend}
\end{table}

\section{Conclusion \& Discussion}

For image editing, existing deep generative models often waste computation by re-synthesizing the image regions that do not require modifications. To solve this issue, we have presented a general-purpose method, \method (\methodabbr), to selectively perform computation on edited regions, and \engine (\engineabbr) to convert the computation reduction to latency reduction on commonly-used hardware. We have demonstrated the effectiveness of our approach in various hardware settings.

\myparagraph{Limitations.}
As discussed in \sect{Ablation Study}, our method requires extra memory to store the original activations, which slightly increases the peak GPU memory usage. It may not work on certain memory-constrained devices, especially for the diffusion models (\eg, DDPM~\cite{song2020denoising}), since our method requires storing activations of all denoising steps. However, recent advancements of few-step samplers~\cite{song2020denoising,xiao2022DDGAN,lu2022dpm,lu2022dpm++,watson2021learning,kong2021fast,meng2022distillation} and low-precision inference~\cite{shang2022post,li2023q} have lowered the memory threshold and make it possible to apply our method to diffusion models.

\looseness=-1
We assume the text input is fixed when applying \engineabbr to Stable Diffusion~\cite{rombach2022high}. For text changes, our method requires recomputing the cached activations, potentially limiting its use cases. Besides, our current method cannot handle text-to-image generation, since even minor adjustments to the text can lead to significant global changes. We leave this as future work. 

\looseness=-1
Our engine has limited speedup on convolutions with low resolution. When the input resolution is low, the active block size needs to be even smaller to get a decent sparsity, such as 1 or 2. However, such extremely small block sizes have bad memory locality and will result in low hardware efficiency. 

\looseness=-1
Besides, we sometimes observe some noticeable boundaries between the edited regions and unedited regions in our generated samples of GauGAN~\cite{park2019semantic}. This is because, for GauGAN model, the unedited regions also change slightly when we perform normal inference. However, since our method does not update the unedited region, there may be some visible seams between the edited and unedited areas, even though the semantics are coherent. Dilating the difference mask would help reduce the gap.

In most cases, the edit will only update the edited regions. However, sometimes the edit will also introduce global illumination changes such as shadow and reflection. For this case, as we only update the edited areas, we cannot update the global changes outside them accordingly.

\looseness=-1 
\myparagraph{Societal impact.}
In this paper, we investigate how to update user edits locally without losing global coherence to enable smoother interaction with the generative models. In real-world scenarios, people could use an interactive interface to edit an image, and our method could provide a quick and high-quality preview for their edits, which eases the process of visual content creation and reduces energy consumption, leading to a greener AI application. The reduced cost also provides a good user experience for lower-end devices, which further democratizes the applications of generative models.

\looseness=-1 However, our method can be utilized by malicious users to generate fake content, deceive people, and spread misinformation, which may lead to potential negative social impacts. Following previous works~\cite{meng2022sdedit}, we explicitly specify the usage permission of our engine with proper licenses. 
Additionally, we run a forensics detector~\cite{wang2020cnn} to detect the generated results of our method. On GauGAN, our generated images can be detected with $97.2\%$ average precision (AP). However, on DDPM~\cite{ho2020denoising,song2020denoising} and Progressive Distillation~\cite{salimans2021progressive}, the APs are only $56.6\%$ and $52.4\%$. Such low APs are caused by the model differences between GANs and diffusion models, as observed in SDEdit~\cite{meng2022sdedit}. We believe developing forensic methods for diffusion models is a critical future research direction.

\ifCLASSOPTIONcompsoc
  \section*{Acknowledgments}
\else
  \section*{Acknowledgment}
\fi

We thank Yaoyao Ding, Zihao Ye, Lianmin Zheng, Haotian Tang, and Ligeng Zhu for the helpful comments on the engine design. We also thank George Cazenavette, Kangle Deng, Ruihan Gao, Daohan Lu, Sheng-Yu Wang, and Bingliang Zhang for their valuable feedback. The project is partly supported by NSF, MIT-IBM Watson AI Lab, Kwai Inc, and Sony Corporation.

\ifCLASSOPTIONcaptionsoff
  \newpage
\fi

\bibliographystyle{IEEEtran}
\bibliography{main}

\clearpage
\clearpage
\appendices

\section{Kernel Fusion}
\lblapp{Kernel Fusion}

\begin{figure*}[t]
    \centering
    \includegraphics[width=\linewidth]{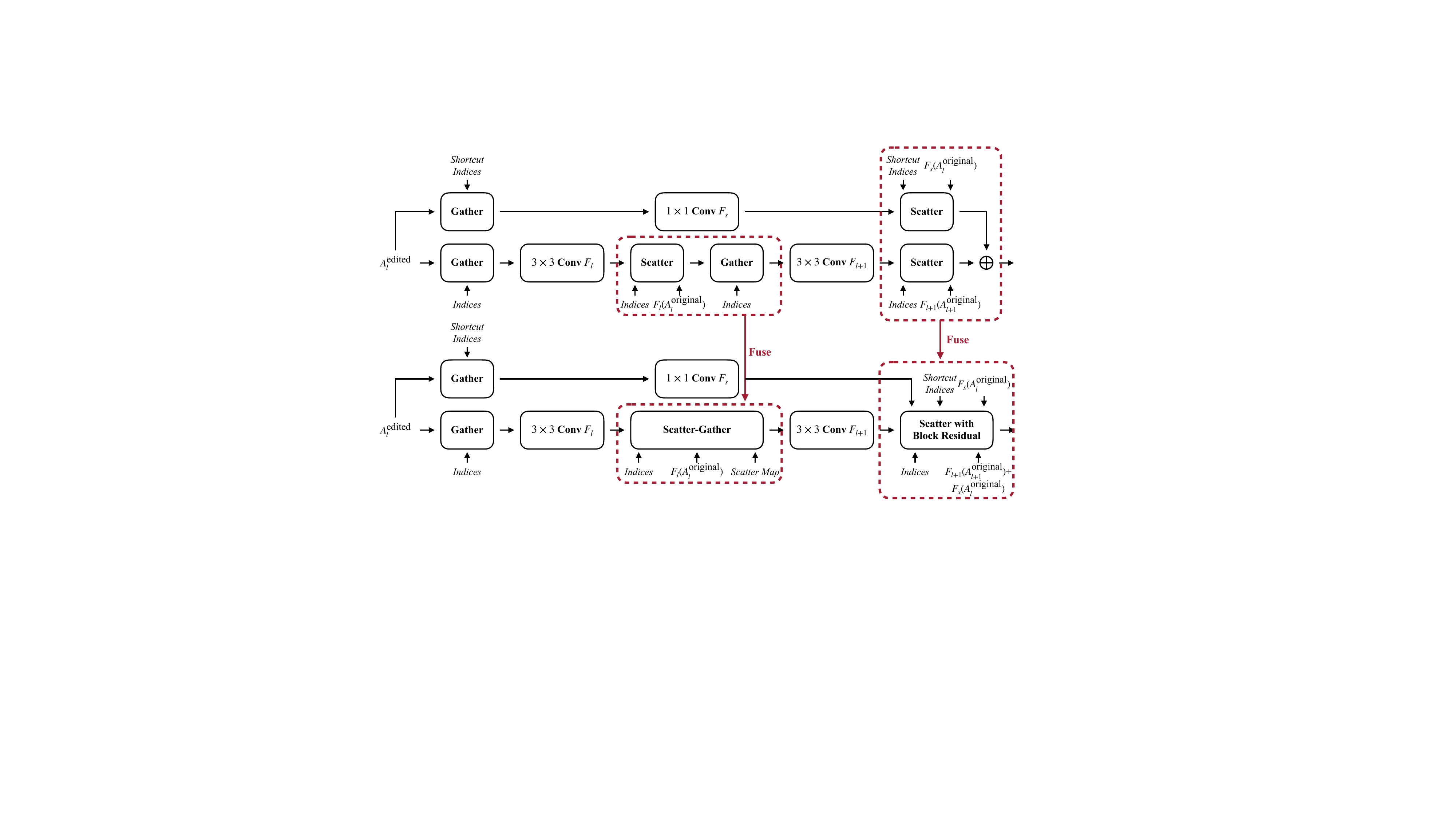}
    \vspace{-15pt}
    \caption{
    	Visualization of kernel fusion in DDPM~\cite{song2020denoising} ResBlock~\cite{he2016deep}. We omit the element-wise operations for simplicity and follow the notations in Section 3. As the kernel sizes of the convolution in the shortcut branch and main branch are different, their reduced active block indices are different (\textit{Indices} and \textit{Shortcut Indices}). To reduce the tensor copying overheads in {\ttfamily Scatter}, we fuse {\ttfamily Scatter} with the following {\ttfamily Gather} into {\ttfamily Scatter-Gather} and fuse the  {\ttfamily Scatter} in the shortcut, main branch, and residual addition into {\ttfamily Scatter with Block Residual}. We pre-compute an additional \textit{Scatter Map} for the {\ttfamily Scatter-Gather} kernel.
    }
    \lblfig{fusion}
\end{figure*}

\begin{figure*}[h]
    \centering
    \includegraphics[width=\linewidth]{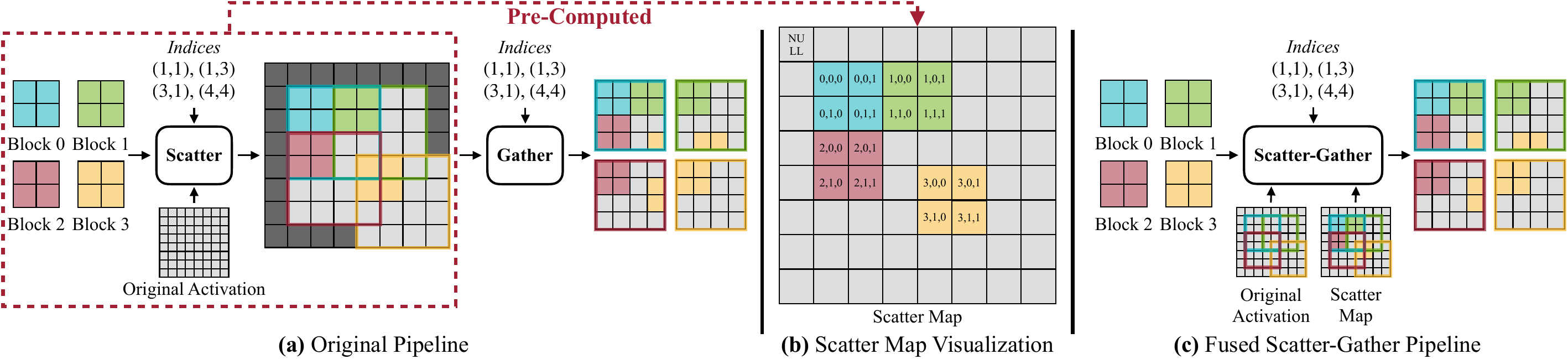}
    \vspace{-15pt}
    \caption{
    Scatter-Gather fusion visualization. \textbf{(a)} The original pipeline of a \texttt{Gather} directly follows a \texttt{Scatter}. The indices indicate the top left corner of the \texttt{Scatter}/\texttt{Gather} position (zero-based). The black blocks are discarded by the {\ttfamily Gather}, which incur redundant data movement. \textbf{(b)} We pre-compute the \texttt{Scatter} process and get a Scatter Map, which tracks the data source during \texttt{Scatter}. If the input data come from the original activation, it stores \texttt{NULL} at this location (gray blocks). Otherwise, it will store a triple locating the data in the input blocks (non-gray blocks). \textbf{(c)} In the fused \texttt{Scatter-Gather} kernel, we directly use the Scatter Map to index and fetch the data from the input blocks and the original activation, avoiding copying the entire original feature map.
    }
    \lblfig{scatter-gather}
\end{figure*}

\looseness=-1 As mentioned in \sect{Sparse Engine}, we fuse {\ttfamily Scatter} and the following {\ttfamily Gather} into a {\ttfamily Scatter-Gather} operator and fuse {\ttfamily Scatter} in the shortcut, main branch, and residual addition together into \texttt{Scatter with Block Residual}. The detailed fusion pattern is shown in \fig{fusion}. For simplicity, we omit the element-wise operations (\eg, {\ttfamily Nonlinearity} and {\ttfamily Scale+Shift}). Below we elaborate on each fusion design. Please refer to our \href{https://github.com/lmxyy/sige}{code} for the detailed implementation.

\subsection{Scatter-Gather Fusion}
\looseness=-1
When a {\ttfamily Gather} directly follows a {\ttfamily Scatter}, we could fuse these two operators into a {\ttfamily Scatter-Gather} to avoid copying the entire original activation $\conv_l(\actori_l)$. As shown in \fig{scatter-gather}(a), in the original pipeline, the black blocks are copied from the original activation and then discarded by {\ttfamily Gather}, which incur redundant data movement. To address this issue, we pre-build a \textit{Scatter Map} to track the data source (\fig{scatter-gather}(b)). For example, if the data at position $(h, w)$ in the \texttt{Scatter} output comes from the original activation, then Scatter Map will store {\ttfamily NULL} at $(h,w)$ (gray blocks). Otherwise, it will store a triple at this position (non-gray blocks). The first element of the triple indicates the block ID that the data come from, while the latter two indicate the offsets of the data within the block.
Note that the pre-computation is cheap and only needs to be computed once for each resolution. Therefore, in the fused \texttt{Scatter-Gather}, we could use the Scatter Map to index and fetch the data directly from either the input blocks or the original activation, given the \texttt{Gather} indices. For example, if we want to fetch the data at location $(h,w)$, we will look up this position in the Scatter Map. If it is \texttt{NULL}, we would fetch the data at location $(h,w)$ in the original activation. Otherwise, we will fetch data in the input blocks indexed by the triple. In this way, we could avoid copying the unused regions in \texttt{Scatter}.

\begin{figure*}[h]
    \centering
    \includegraphics[width=\linewidth]{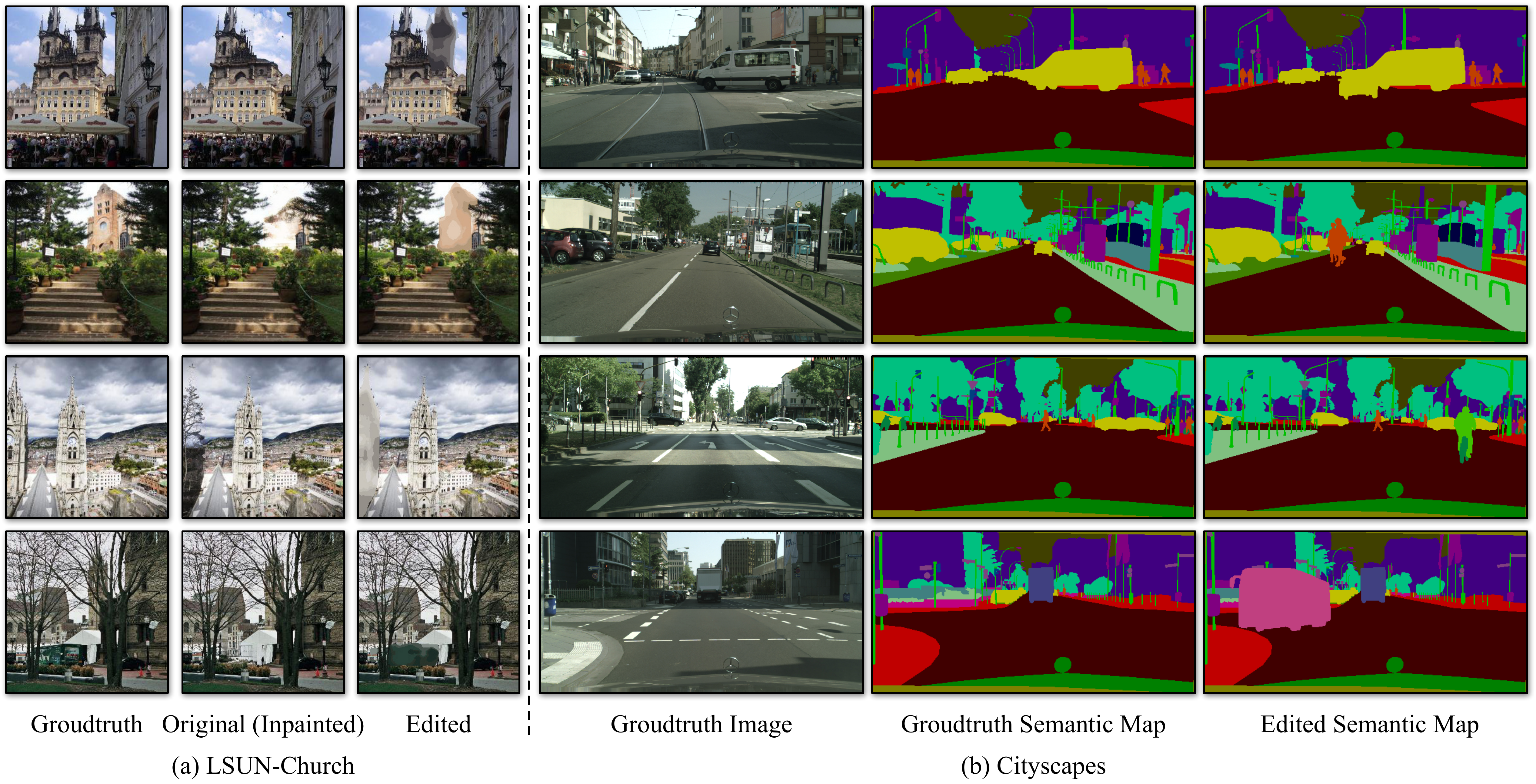}
    \vspace{-20pt}
    \caption{
    \looseness=-1
    Examples of our synthetic edits on (a) LSUN Church and (b) Cityscapes. On LSUN Church, we view the inpainted image as the original image and generate the edits by quantizing color at the corresponding regions. On Cityscapes, we generate the edits by pasting some foreground objects on the ground-truth semantic maps.}
    \lblfig{dataset}
\end{figure*}
\begin{figure*}[h]
\captionsetup[subfigure]{font=scriptsize}
\centering
\subfloat[LSUN Church.\label{fig:church-editing}]{
	\includegraphics[width=0.485\textwidth]{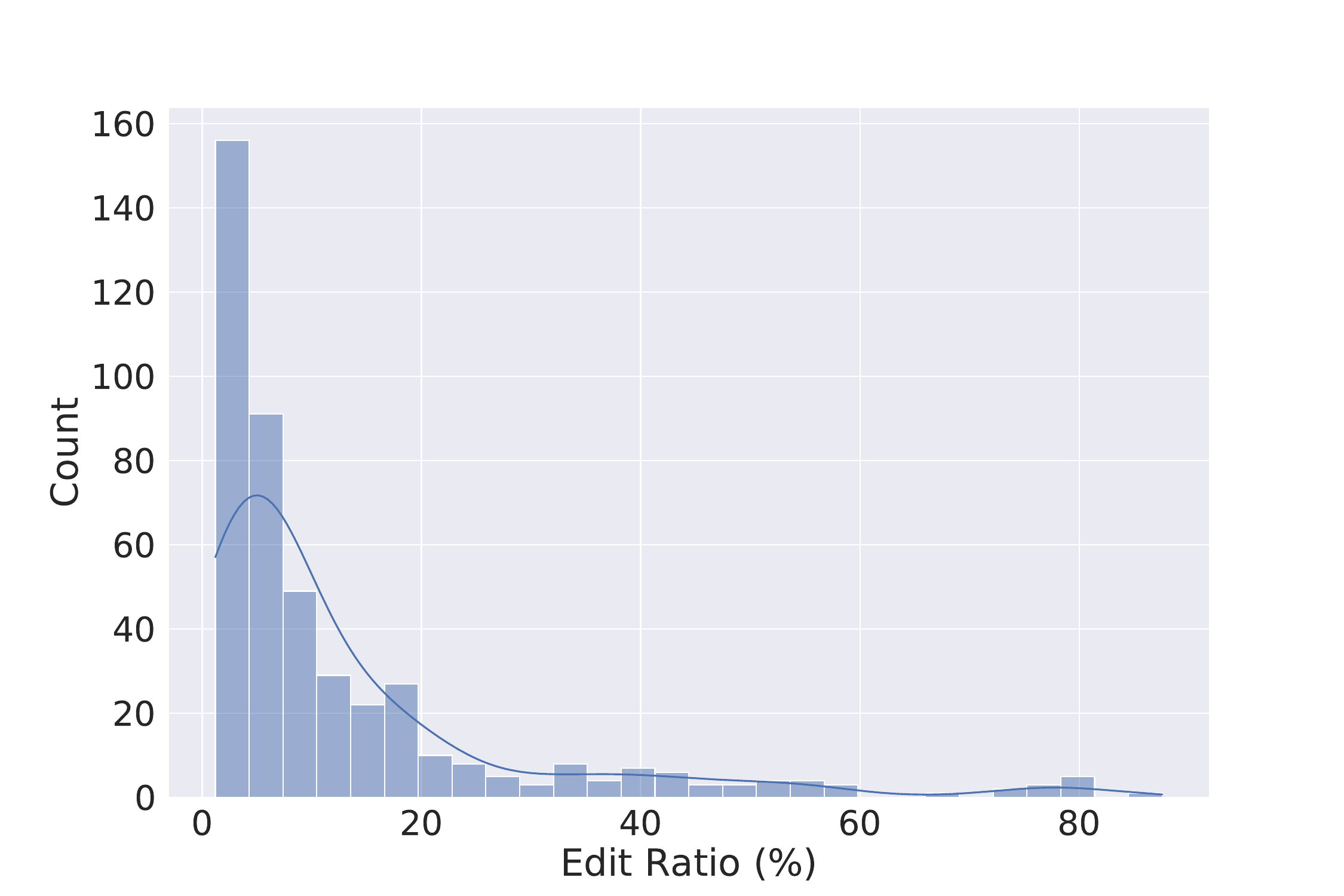}
}
\hfill
\subfloat[Cityscapes.\label{fig:cityscapes-editing}]{
    \includegraphics[width=0.485\textwidth]{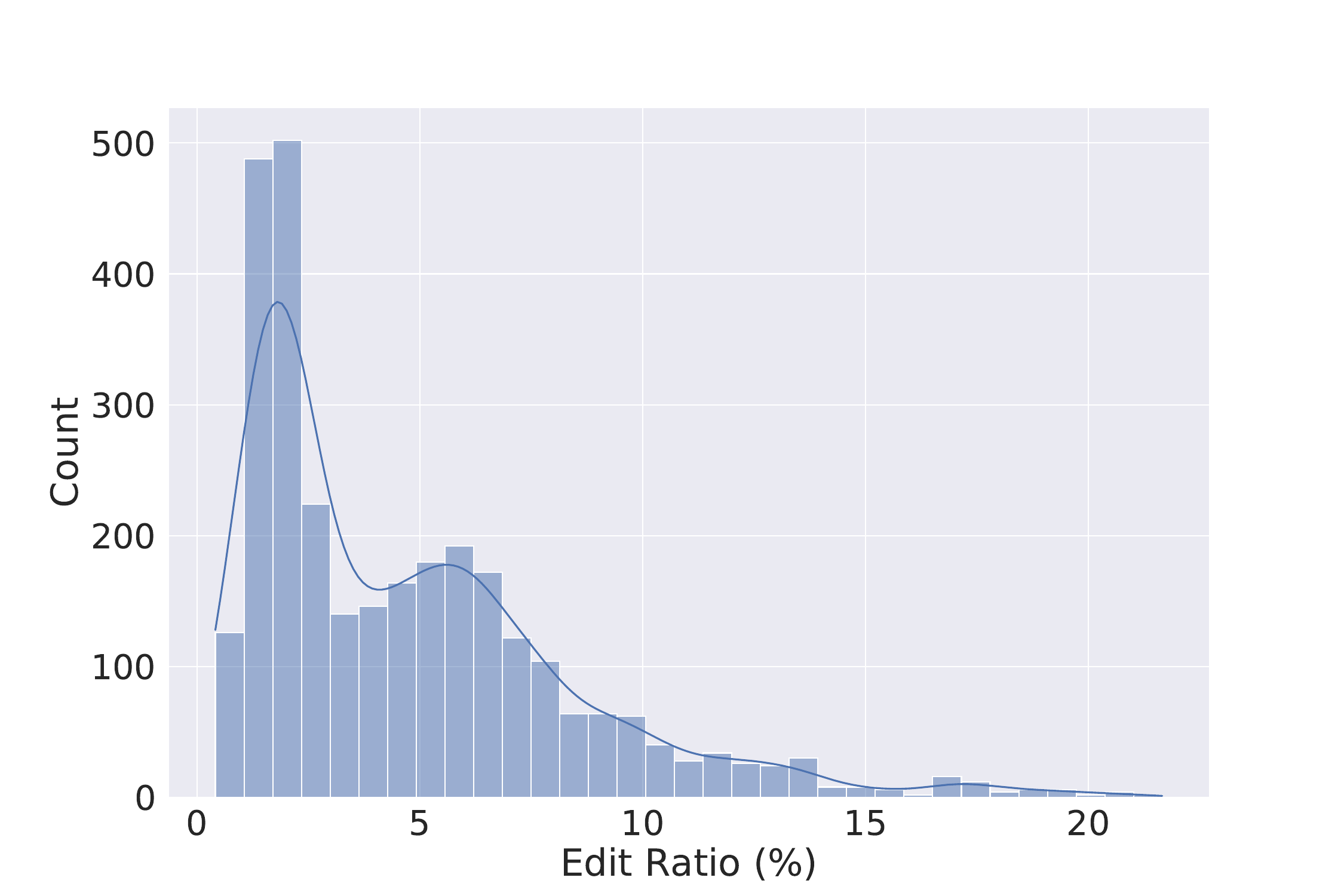}
}
\vspace{-5pt}
\caption{Detailed edit ratio distribution of our synthetic datasets.}
\end{figure*}

\subsection{Shortcut Scatter Fusion}
\looseness=-1
The $1 \times 1$ convolution in the shortcut branch consumes much less computation than the convolution in the main branch. Therefore, the overheads of {\ttfamily Gather} and {\ttfamily Scatter} weigh more in the shortcut branch. We fuse the {\ttfamily Scatter} in the shortcut branch and main branch along with residual addition into {\ttfamily Scatter with Block Residual} to reduce these overheads. Specifically, as shown in \fig{fusion}, we first scatter $\conv_{l+1}$ output into the pre-computed $\conv_{l+1}(\actori_l)+\conv_s(\actori_l)$ and add the original residual $\conv_s(\actori_l)$ only at the scattered locations correspondingly according to \textit{Indices}. Then we calibrate the resulting feature map with $\conv_s$ output by adding the residual difference $\conv_s(\actedi_l)-\conv_s(\actori_l)$ at the scattered locations indexed by \textit{Shortcut Indices} in place. 

\section{Benchmark Datasets}
\lblapp{Benchmark Datasets}
\looseness=-1 Below, we elaborate on how we build the synthetic datasets.

\myparagraph{LSUN Church.} \fig{dataset}(a) shows some examples of our synthetic edits on LSUN Church. The average edited area of the whole dataset is 13.1\%. The detailed distribution is shown in \fig{church-editing}.

\myparagraph{Cityscapes.} We collect 27 foreground object semantic masks from the validation set. The objects include 4 bicycles, 1 motorcycle, 7 cars, 6 trucks, 3 buses, 5 persons, and 1 train. \fig{masks}(a) visualizes some collected semantic masks. We generate the edits by randomly pasting one of these objects to the ground-truth semantic maps with augmentation. The augmentation includes random horizontal flip, resize (scale factor in $[0.8, 1.2]$), and translation ($[-32, 32]$ for vertical one and $[-64,64]$ for horizontal one). To make the synthetic edits more reasonable, when the scale factor is larger than 1, the vertical translation can only be positive. Otherwise, it can only be negative. \fig{dataset}(b) shows some edit examples. The average edited area of the entire dataset is 4.77\%. The detailed distribution is shown in \fig{cityscapes-editing}.

\myparagraph{LAION.} We automatically construct our inpainting masks by overlaying circles of random sizes at arbitrary positions. Additional mask examples are displayed in \fig{masks}(b).

\begin{figure*}[t]
    \centering
    \includegraphics[width=\linewidth]{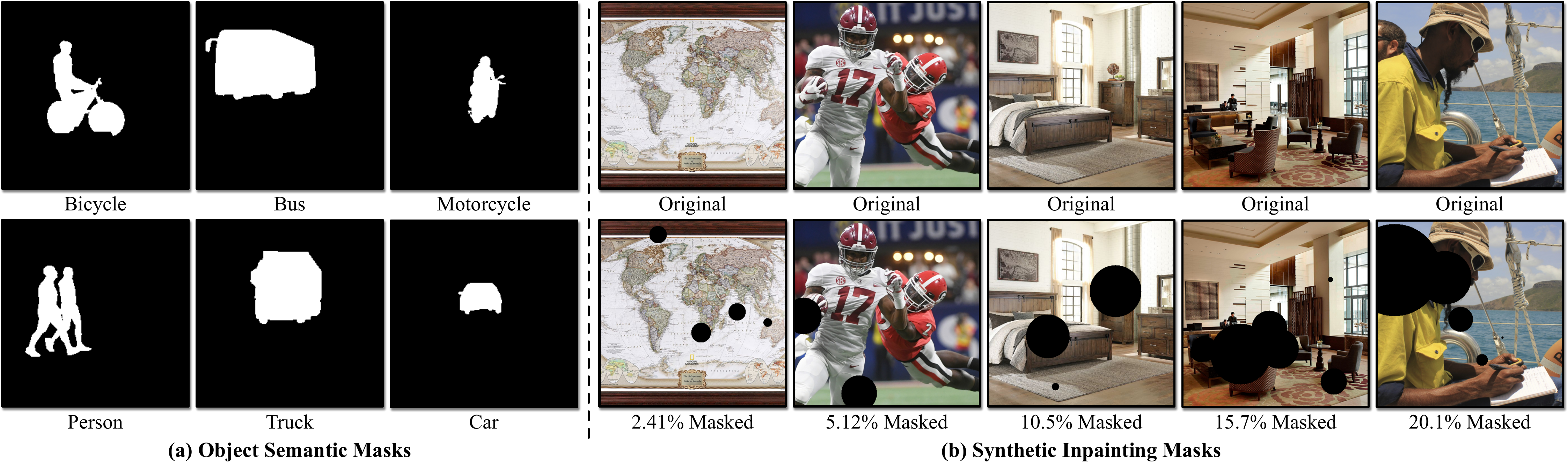}
    \vspace{-20pt}
    \caption{(a) Examples of collected foreground object semantic masks on Cityscapes~\cite{cordts2016cityscapes}. (b) Mask examples of the synthetic inpainting datasets on LAION-5B~\cite{schuhmann2022laion}.}
	\lblfig{masks}
	\vspace{-15pt}
\end{figure*}
\renewcommand \arraystretch{1.}
\begin{table}[t]
\setlength{\tabcolsep}{1.45pt}
\scriptsize \centering
\begin{tabular}{lccccccc}
\toprule
 \multirow{2}{*}[\multirowcenter]{Method} &\multicolumn{2}{c}{MACs} & \multicolumn{2}{c}{PSNR ($\uparrow$)} & \multicolumn{2}{c}{LPIPS ($\downarrow$)}  & \multirow{2}{*}[\multirowcenter]{mIoU ($\uparrow$)} \\
\cmidrule(lr){2-3} \cmidrule(lr){4-5} \cmidrule(lr){6-7}
& Value & Ratio & w/ G.T. & w/ Orig. & w/ G.T. & w/ Orig. \\
\midrule
Original & 281G & -- & 15.9 & -- & 0.414 & -- & 57.3 \\
\cmidrule{1-8}
GAN Comp.~\cite{li2020gan} & 31.2G & 9.0$\times$ & 15.8 & 19.1 & \textbf{0.417} & 0.329 & \textbf{56.3} \\
\cmidrule{1-8}
\textbf{Ours} & \textbf{30.7G} & \textbf{9.2}$\times$ & \textbf{15.9} & \textbf{27.5} & 0.425 & \textbf{0.076} & 56.1 \\
\cmidrule[0.9pt]{1-8}
0.19 GauGAN & 13.3G & 21$\times$ & 15.4 & 18.4 & 0.427 & 0.356 & 49.5 \\
 \cmidrule{1-8}
GAN Comp. (S) & 9.64G & 29$\times$ & 15.8 & \textbf{18.9} & \textbf{0.422} & \textbf{0.344} & 51.2 \\
\cmidrule{1-8}
\textbf{GAN Comp.+Ours} & \textbf{7.06G} & \textbf{40}$\times$ & \textbf{15.8} & 18.8 & 0.429 & 0.345 &  \textbf{52.4}  \\
\bottomrule
\end{tabular}
\vspace{-5pt}
\caption{
\looseness=-1
Quality evaluation of GauGAN at the edited regions. PSNR/LPIPS \textit{w/ G.T.} means computing the metrics with the ground-truth images, and \textit{w/ Orig.} means computing with the generated samples from the original model. \textit{0.19 GauGAN}: Reducing each layer of GauGAN to 19\% channels and training from scratch. \textit{GAN Comp. (S)}: GAN Compression with larger compression ratio. Our method matches the performance of GAN Compression~\cite{li2020gan}. When applying it to GAN Compression, our method demonstrates comparable results to \textit{GAN Comp. (S)} with less computation, achieving a $40\times$ MACs reduction.
}
\lbltab{quality-edited}
\end{table}

\section{Additional Results}

\myparagraph{Quality results at the edited regions.} 
In \tab{quality}, we show the quantitative quality results of our method. For DDPM and PD, the unedited areas in the generated images keep the same as the input images due to the mask trick in SDEdit~\cite{meng2022sdedit}. For GauGAN, the generated unedited regions vary across different methods. In this case, the image quality in these areas will influence the metrics we report in \tab{quality}. We also include the quantitative quality results of GauGAN at the edited regions in \tab{quality-edited}. Our method could still preserve the image quality of the original GauGAN and match the performance of GAN Compression~\cite{li2020gan}. When applied to GAN Compression, it reduces $40\times$ MACs on average, achieving results on par with \textit{GAN Comp. (S)} with less computation.

\myparagraph{Additional visualization.}
In \fig{res-church}, we show more visual results of DDPM~\cite{song2020denoising} and Progressive Distillation~\cite{salimans2021progressive} on LSUN Church~\cite{yu15lsun}. Besides LSUN Church, we additionally apply \engineabbr to DDPM on CelebA~\cite{liu2015faceattributes} and AFHQ Dogs\footnote{The pre-trained models on CelebA and AFHQ dogs are from \href{https://github.com/ermongroup/SDEdit}{SDEdit}~\cite{meng2022sdedit} and \href{https://github.com/jychoi118/ilvr_adm}{ILVR}~\cite{choi2021ilvr}, respectively.}~\cite{choi2020starganv2}, as shown in \fig{celeba_afhq}, respectively. In \fig{res-cityscapes}, we show more visual results of GauGAN on Cityscapes~\cite{cordts2016cityscapes}. In \fig{res-stable-diffusion}, we show more visual results of Stable Diffusion~\cite{rombach2022high} on image-to-image translation.

\section{License \& Computation Resources}
Here we show all the licenses of our used assets. 
Our backbone models \href{https://github.com/ermongroup/ddim}{DDIM}~\cite{song2020denoising}, \href{https://github.com/jychoi118/ilvr_adm}{ILVR}~\cite{choi2021ilvr}, \href{https://github.com/google-research/google-research/tree/master/diffusion_distillation}{Progressive Distillation}~\cite{salimans2021progressive}, \href{https://github.com/CompVis/stable-diffusion}{Stable Diffusion}, \href{https://github.com/NVlabs/SPADE}{GauGAN}~\cite{park2019semantic} and \href{https://github.com/mit-han-lab/gan-compression}{GAN Compression}~\cite{li2020gan} is under \href{https://github.com/ermongroup/ddim/blob/main/LICENSE}{MIT license}, \href{https://github.com/jychoi118/ilvr_adm/blob/main/LICENSE}{MIT license}, \href{https://github.com/google-research/google-research/blob/master/LICENSE}{Apache license}, \href{https://github.com/CompVis/stable-diffusion/blob/main/LICENSE}{CreativeML Open RAIL-M}, \href{https://github.com/NVlabs/SPADE/blob/master/LICENSE.md}{Creative Commons license} and \href{https://github.com/mit-han-lab/gan-compression/blob/master/LICENSE}{BSD license}, respectively.
\href{https://github.com/ermongroup/SDEdit}{SDEdit} is under \href{https://github.com/ermongroup/SDEdit/blob/main/LICENSE}{MIT license}. 
The license of Cityscapes~\cite{cordts2016cityscapes} is \href{https://www.cityscapes-dataset.com}{here}. The CelebA~\cite{liu2015faceattributes} license is \href{https://mmlab.ie.cuhk.edu.hk/projects/CelebA.html}{here}. AFHQ-Dogs~\cite{choi2020starganv2} is under \href{https://github.com/clovaai/stargan-v2/blob/master/LICENSE}{Creative Common license}. LSUN Church~\cite{yu15lsun} does not have an explicit license. The examples in \fig{stable-diffusion}(a) and \fig{res-stable-diffusion} are under Creative Commons License. The drawing in \fig{stable-diffusion}(b) is created by ourselves, referring to the painting in this \href{https://www.mooyuu.com/uploadfile/2019/1108/20191108063502962.jpg}{link}.

Since our method does not involve any model training, all our generated results are obtained on a single NVIDIA RTX 3090, which only takes $1 \sim 2$ hours to process all the test images ($\sim 10,000$ in total), including both the original models and our method. We measure the model latency on NVIDIA RTX 3090, 2080Ti, Intel Core i9-10920X CPU, and Apple M1 Pro CPU. On Apple M1 Pro CPU, we use Intel Anaconda for our Python environment.

\section{Changelog}
\myparagraph{V1} Initial preprint release (NeurIPS 2022).

\myparagraph{V2} Fix the figure display issue on mobile devices. 

\myparagraph{V3} (a) Add additional method details. (b) Support attention layers and add the results of Stable Diffusion~\cite{rombach2022high} (\sect{Extension to attention layers.} and \fig{quality}). 

\myparagraph{V4} Accepted by T-PAMI. (a) Support MPS backend (\tab{efficiency}). (b) Curate an inpainting benchmark from LAION-5B~\cite{schuhmann2022laion} to evaluate Stable Diffusion~\cite{rombach2022high} (\tab{quality} and \fig{quality}). (c) Expanded large editing analysis (\fig{curves}). (d) Additional visual results on multiple datasets (\figs{celeba_afhq} and \ref{fig:res-stable-diffusion}).

\begin{figure*}[t]
    \centering
    \includegraphics[width=\linewidth]{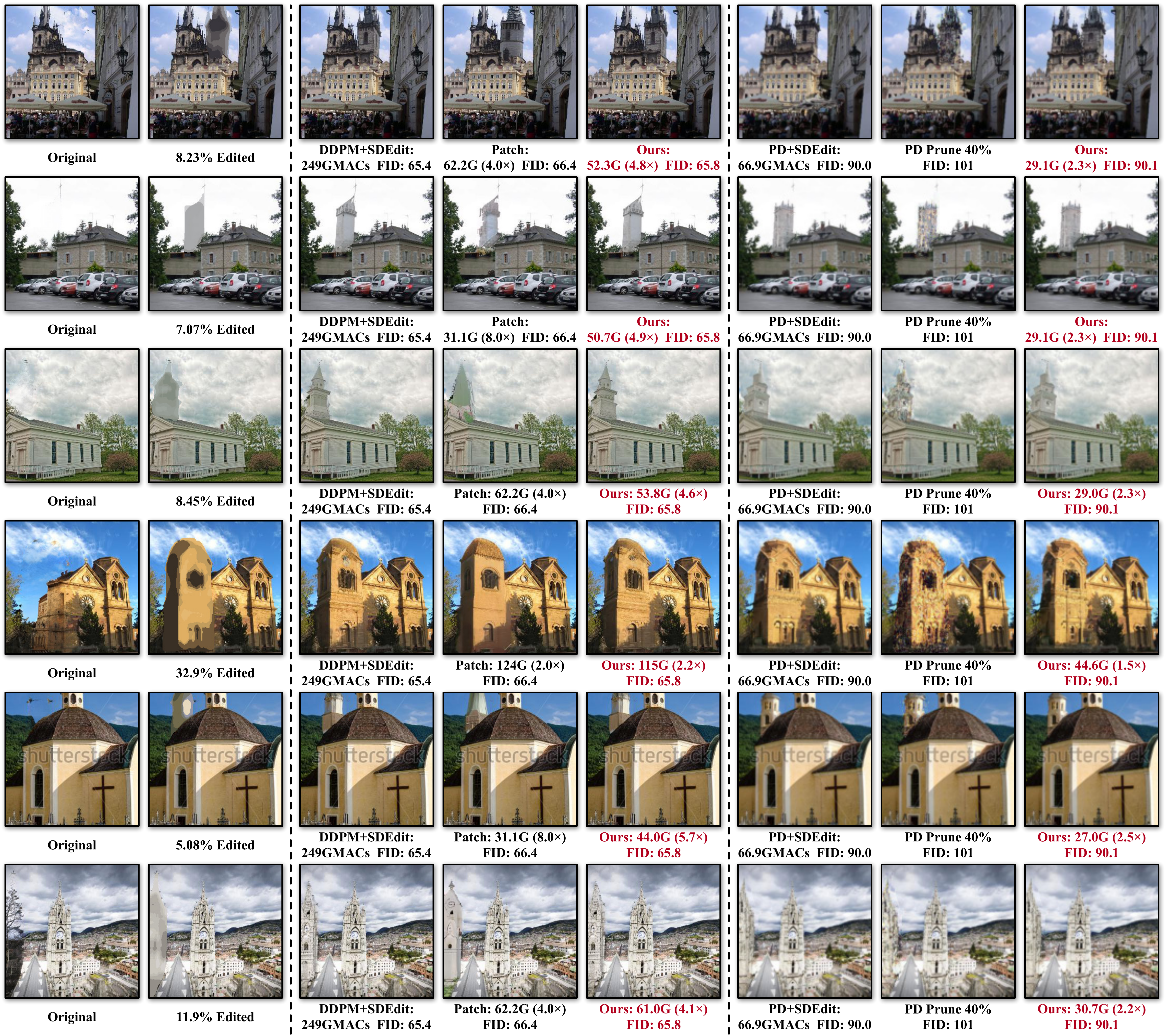}
    \vspace{-20pt}
    \caption{More visual results on LSUN Church of SDEdit with DDPM~\cite{song2020denoising} and Progressive Distillation. MACs measure the computation for a single model forward. \textit{Prune 40\%}: Uniformly pruning 40\% weights of the model without fine-tuning. \textit{Patch}: Cropping the smallest image patch that covers all the edited regions of the model input and blending the model output back to the input image. Our method achieves lower FID with fewer MACs for both DDPM and Progressive Distillation.}
    \lblfig{res-church}
\end{figure*}
\begin{figure*}[t]
    \centering
    \includegraphics[width=\linewidth]{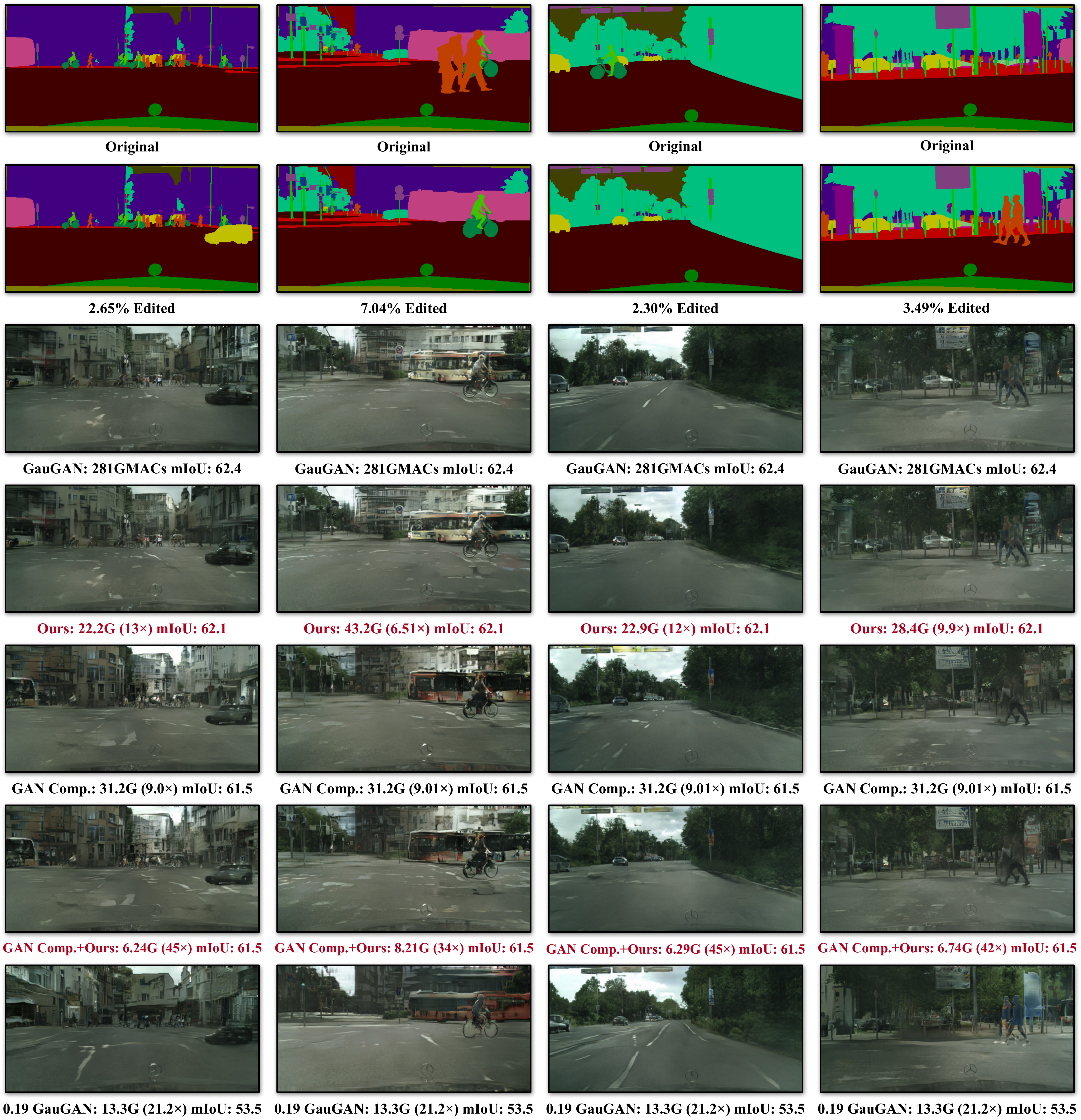}
    \vspace{-20pt}
    \caption{More visual results on Cityscapes of GauGAN~\cite{park2019semantic}. \textit{0.19 GauGAN}: Uniformly reducing each layer of GauGAN to 19\% channels and training from scratch. Our method could achieve higher mIoU than GAN Compression with fewer MACs. When applying to GAN Compression, our method achieves a $34\sim45 \times$ MACs reduction with a minor mIoU drop.}
    \lblfig{res-cityscapes}
\end{figure*}
\begin{figure*}[h]
    \centering
    \includegraphics[width=\linewidth]{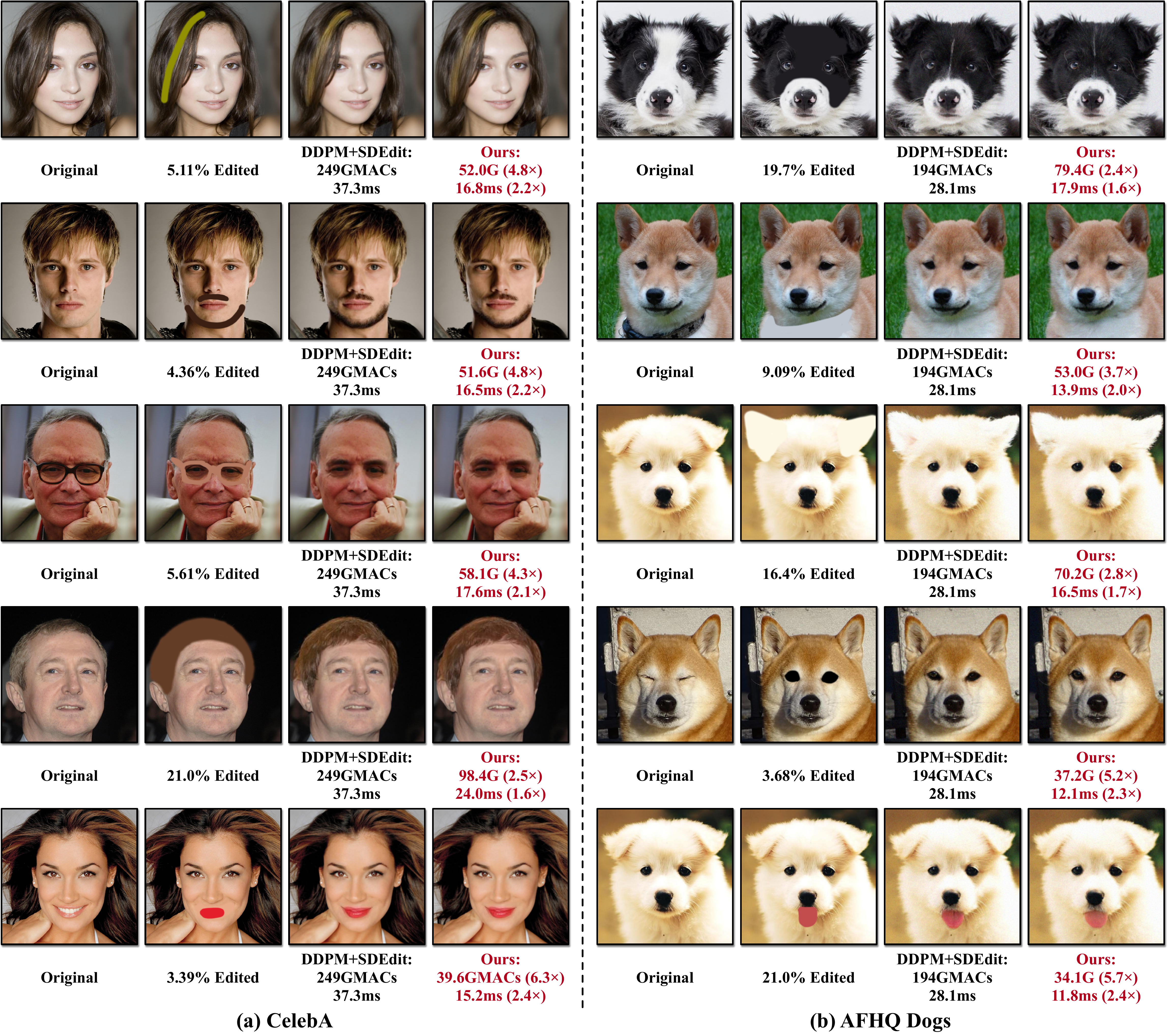}
    \vspace{-20pt}
    \caption{\looseness=-1 Visual results of DDPM~\cite{ho2020denoising} on CelebA~\cite{liu2015faceattributes} and AFHQ Dogs~\cite{choi2020starganv2}. The computation and latency are measured for a single diffusion step on NVIDIA RTX 3090. \engineabbr reduces the computation of the original by up to $6.3\times$, achieving an up to $2.4\times$ speedup.}
    \lblfig{celeba_afhq}
\end{figure*}
\begin{figure*}[h]
    \centering
   \includegraphics[width=\linewidth]{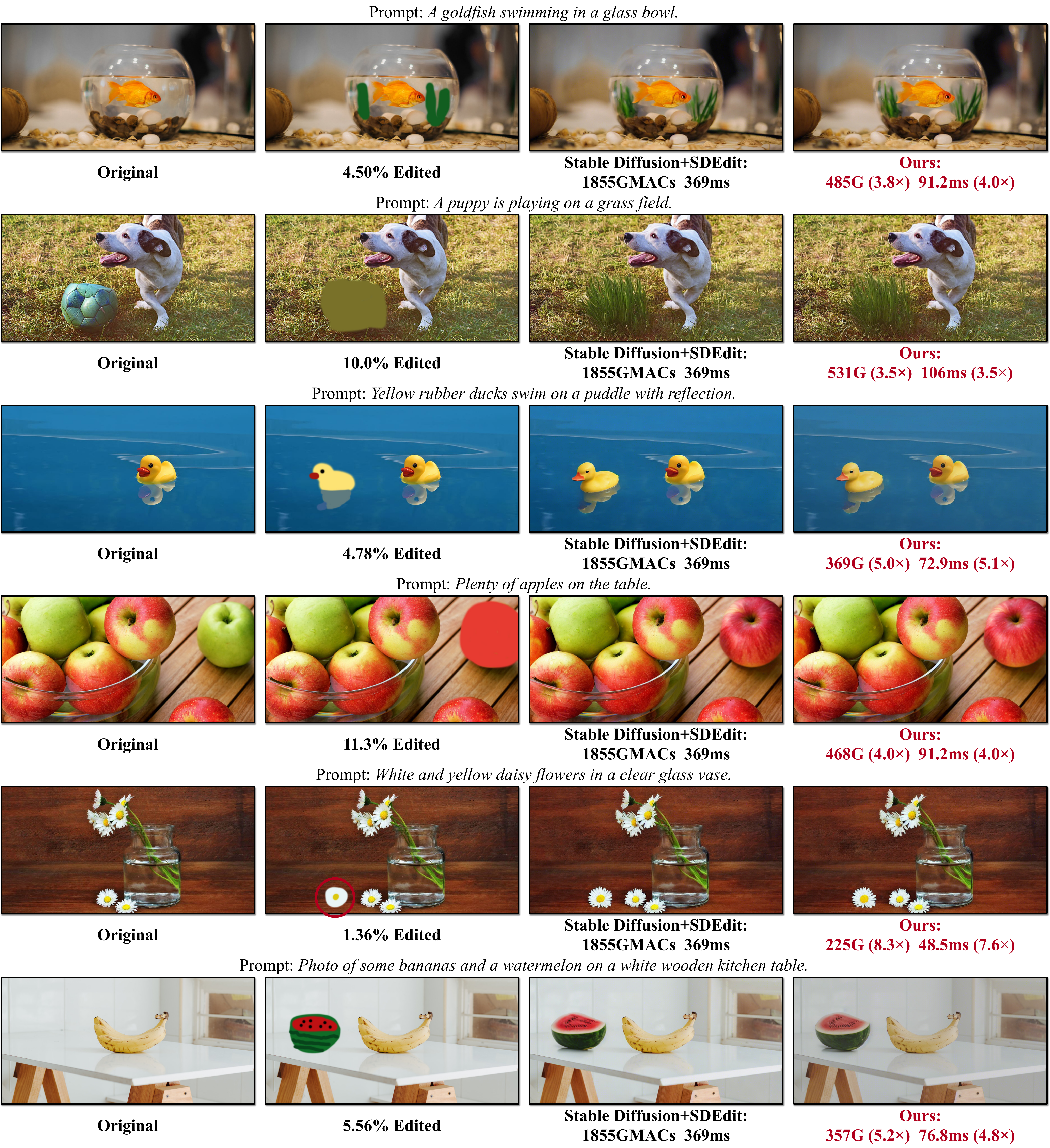}
    \caption{Visual results of Stable Diffusion on text-guided image editing. The computation and latency are measured for a single diffusion step on NVIDIA RTX 3090. \engineabbr achieves an up to $7.6\times$ speedup under a $1.4\%$ edit.}
    \lblfig{res-stable-diffusion}
\end{figure*}

\end{document}